\documentclass[Afour,sageh,times]{sagej}

\usepackage{moreverb,url}

\usepackage[colorlinks,bookmarksopen,bookmarksnumbered,citecolor=red,urlcolor=red]{hyperref}

\newcommand\BibTeX{{\rmfamily B\kern-.05em \textsc{i\kern-.025em b}\kern-.08em
T\kern-.1667em\lower.7ex\hbox{E}\kern-.125emX}}

\def\volumeyear{2022}

\usepackage{multicol}
\usepackage{float}
\usepackage{amsfonts}
\usepackage{comment}
\usepackage{amsmath}
\usepackage{changepage}
\usepackage{amssymb}
\usepackage{graphicx}
\usepackage{adjustbox}
\usepackage{latexsym}
\usepackage{enumerate}
\usepackage[font=small,belowskip=0pt]{caption}
\usepackage{mathtools}
\usepackage{algorithm}
\usepackage{algcompatible}
\usepackage{algpseudocode}
\usepackage{bbm}
\usepackage{xcolor}
\usepackage{bm}
\usepackage{booktabs}
\usepackage[shortlabels]{enumitem}
\usepackage{subfig}
\usepackage{array}
\usepackage{wasysym}

\usepackage{nomencl}
\makenomenclature

\renewcommand{\nompreamble}{The list below summarizes important symbols in the paper.}

\usepackage{etoolbox}
\renewcommand\nomgroup[1]{%
  \item[\bfseries
  \ifstrequal{#1}{A}{Interaction planning}{%
  \ifstrequal{#1}{B}{Scenario tree and MPC}{%
  \ifstrequal{#1}{C}{Shielding and SHARP}{}}}%
]}

\newcommand{\upperRomannumeral}[1]{\uppercase\expandafter{\romannumeral#1}}

\definecolor{porange}{HTML}{E77500} 
\algrenewcommand{\algorithmiccomment}[1]{\textcolor{red}{\hfill// #1}}
\algnewcommand{\LineComment}[1]{\Statex \texttt{\textcolor{red}{// #1}}}

\usepackage{amsmath}
\usepackage{bm}

\usepackage{amsthm}
\newtheorem{theorem}{Theorem}
\newtheorem{lemma}{Lemma}    

\newtheorem{remark}{Remark}
\newtheorem{example}{Example}    

\newtheorem{proposition}{Proposition}    
\newtheorem{assumption}{Assumption}    
\newtheorem{definition}{Definition}

\usepackage{ifthen}
\newboolean{include-notes}
\newboolean{include-new}
\newboolean{include-remove}
\setboolean{include-notes}{true}
\setboolean{include-new}{true}
\setboolean{include-remove}{false}

\usepackage{xcolor}
\usepackage[normalem]{ulem}
\definecolor{newcolor}{rgb}{0, 0, 1}
\definecolor{removecolor}{rgb}{1, 0, 0}
\newcommand{\jaime}[1]{\ifthenelse{\boolean{include-notes}}{\textcolor{orange}{\textbf{JFF:} #1}}{}}
\newcommand{\haimin}[1]{\ifthenelse{\boolean{include-notes}}{\textcolor{teal}{\textbf{HH:} #1}}{}}
\newcommand{\david}[1]{\ifthenelse{\boolean{include-notes}}{\textcolor{cyan}{\textbf{David:} #1}}{}}
\newcommand{\sangjae}[1]{\ifthenelse{\boolean{include-notes}}{\textcolor{green}{\textbf{Sangjae:} #1}}{}}
\newcommand{\remove}[1]{\ifthenelse{\boolean{include-remove}}{\textcolor{removecolor}{\sout{#1}}}{}}
\newcommand{\new}[1]{\ifthenelse{\boolean{include-new}}{\textcolor{newcolor}{#1}}{#1}}

\usepackage{lipsum}

\newcommand{\reals}{\mathbb{R}}

\newcommand{\distr}{p}
\newcommand{\prob}{P}
\newcommand{\probbar}{\bar{P}}
\newcommand{\mean}{\mu}

\newcommand{\covar}{\Sigma}
\newcommand{\entropy}{H}
\newcommand{\gaussian}{{\mathcal{N}}}

\newcommand{\ivec}{\mathcal{I}}
\newcommand{\bel}{b}

\DeclareMathOperator*{\expectation}{\mathbb{E}}

\DeclareMathOperator*{\argmax}{arg\,max}

\DeclareMathOperator*{\subjectto}{subject\,\, to \quad}

\newcommand{\diag}{\operatorname{diag}}
\newcommand{\trace}{\operatorname{tr}}

\newcommand{\state}{{x}}
\newcommand{\dstate}{{\delta\state}}
\newcommand{\bstate}{{\bar{\state}}}
\newcommand{\ctrl}{{u}}
\newcommand{\dstb}{{d}}
\newcommand{\dstbbar}{{\bar{d}}}

\newcommand{\cset}{{\mathcal{U}}}
\newcommand{\dset}{{\mathcal{D}}}

\newcommand{\dyn}{{f}}
\newcommand{\beldyn}{{g}}
\newcommand{\beldynapprox}{{\tilde{g}}}

\newcommand{\valfunc}{{V}}
\newcommand{\qfunc}{{Q}}

\newcommand{\policy}{{\pi}}

\newcommand{\failure}{{\mathcal{F}}}
\newcommand{\shield}{s}
\newcommand{\sfilter}{f}
\newcommand{\shieldset}{\mathcal{S}}

\renewcommand\algorithmicfunction{\textbf{function}}
\renewcommand\Function{\item[ \algorithmicfunction]}

\newcommand{\mset}{{\mathcal{M}}}
\newcommand{\ego}{{e}}
\newcommand{\oppo}{{o}}

\newcommand{\nodeset}{\mathcal{N}}
\newcommand{\nodesetleaf}{\mathcal{L}}
\newcommand{\cnodeset}{\mathcal{C}}

\newcommand{\node}{n}
\newcommand{\tnode}{{\tilde{\node}}}
\newcommand{\pre}[1]{\operatorname{\mathfrak{p}}(#1)}

\newcommand{\cl}{\text{cl}}
\newcommand{\Sim}{\text{sim}}
\newcommand{\yes}{\textbf{Y}}
\newcommand{\no}{N}

\begin{document}

\runninghead{Hu et al.}

\title{Active Uncertainty Reduction for Safe and Efficient Interaction Planning: A Shielding-Aware Dual Control Approach}

\author{Haimin Hu\affilnum{1}, David Isele\affilnum{2}, Sangjae Bae\affilnum{2}, and Jaime F. Fisac\affilnum{1}}

\affiliation{\affilnum{1}Department of Electrical and Computer Engineering, Princeton, Princeton University, NJ 08544, USA\\
\affilnum{2}Honda Research Institute, San Jose, CA 95134, USA}

\corrauth{Haimin Hu, Department of Electrical and Computer Engineering, Princeton University, Princeton, NJ 08544, USA}

\email{haiminh@princeton.edu}

\begin{abstract}
The ability to accurately predict others' behavior is central to the safety and efficiency of robotic systems in interactive settings, such as human-robot interaction and multi-robot teaming tasks.
Unfortunately, robots often lack access to key information on which these predictions may hinge, such as other agents' goals, attention, and willingness to cooperate.
Dual control theory addresses this challenge by treating unknown parameters of a predictive model as stochastic hidden states and inferring their values at runtime using information gathered during system operation.
While able to optimally and automatically trade off exploration and exploitation, dual control is computationally intractable for general interactive motion planning, mainly due to the fundamental coupling between the robot's trajectory plan and its prediction of other agents' intent.
In this paper, we present a novel algorithmic approach to enable active uncertainty reduction for interactive motion planning based on the implicit dual control paradigm.
Our approach relies on sampling-based approximation of stochastic dynamic programming, leading to a model predictive control problem that can be readily solved by real-time gradient-based optimization methods.
The resulting policy is shown to preserve the dual control effect for a broad class of predictive models with both continuous and categorical uncertainty.
To ensure the safe operation of the interacting agents, we use a runtime safety filter (also referred to as a ``shielding'' scheme), which overrides the robot's dual control policy with a safety fallback strategy when a safety-critical event is imminent.
We then augment the dual control framework with an improved variant of the recently proposed shielding-aware robust planning scheme, which proactively balances the nominal planning performance with the risk of high-cost emergency maneuvers triggered by low-probability agent behaviors.
We demonstrate the efficacy of our approach with both simulated driving studies and hardware experiments using 1/10 scale autonomous vehicles. 
\end{abstract}


\maketitle

\section{Introduction}
Computing robot plans that account for possible interactions with one or multiple other agents is a challenging task, as the robotic system and other agents may have coupled dynamics, limited communication capabilities, and conflicting interests.
Examples of interaction planning under uncertainty include human-robot interaction~\cite{fisac2018general,sadigh2018planning,bajcsy2021analyzing}, multi-robot teaming~\cite{leonard2007collective,tokekar2016sensor,santos2018coverage}, swarm robotics~\cite{swain2011real,rubenstein2014programmable,chung2018survey}, autonomous racing~\cite{liniger2015optimization,kabzan2019learning,schwarting2021stochastic}, and mixed-autonomy traffic~\cite{isele2019interactive,bae2020cooperation,wu2021flow}.
To achieve safety and efficiency in those scenarios, the robot must competently predict and seamlessly adapt to the other agent's behavior.
Intent-driven behavior models are widely used for such predictions: for example, the Boltzmann model~\cite{luce1959individual,ziebart2008maximum} is commonly used for motion prediction of noisily rational decision makers.
This model assumes that the other agent is exponentially more likely to take actions with a higher underlying \emph{utility}.
If the other agent's intent is well captured by a given utility function, the interaction can be modeled as a dynamic game in which the players' feedback strategies can be obtained via dynamic programming~\cite{fisac2019hierarchical}.
However, typical interaction settings may admit a plethora of \emph{a priori} plausible intents (e.g., corresponding to distinct equilibrium solutions~\cite{peters2020inference} or different agent's preferences~\cite{sadigh2018planning}), which in general cannot be fully modeled, let alone observed, by the robot~\cite{fisacBHFWTD18}.
The robot may seek to represent the other agent's intent through a parametric model and then infer the value of these parameters as hidden states under a Bayesian framework~\cite{fisacBHFWTD18,tian2021safety,hu2022sharp}, but doing so tractably while planning through interactions is an open problem.

\begin{figure*}[!hbtp]
  \centering
  \includegraphics[width=1.95\columnwidth]{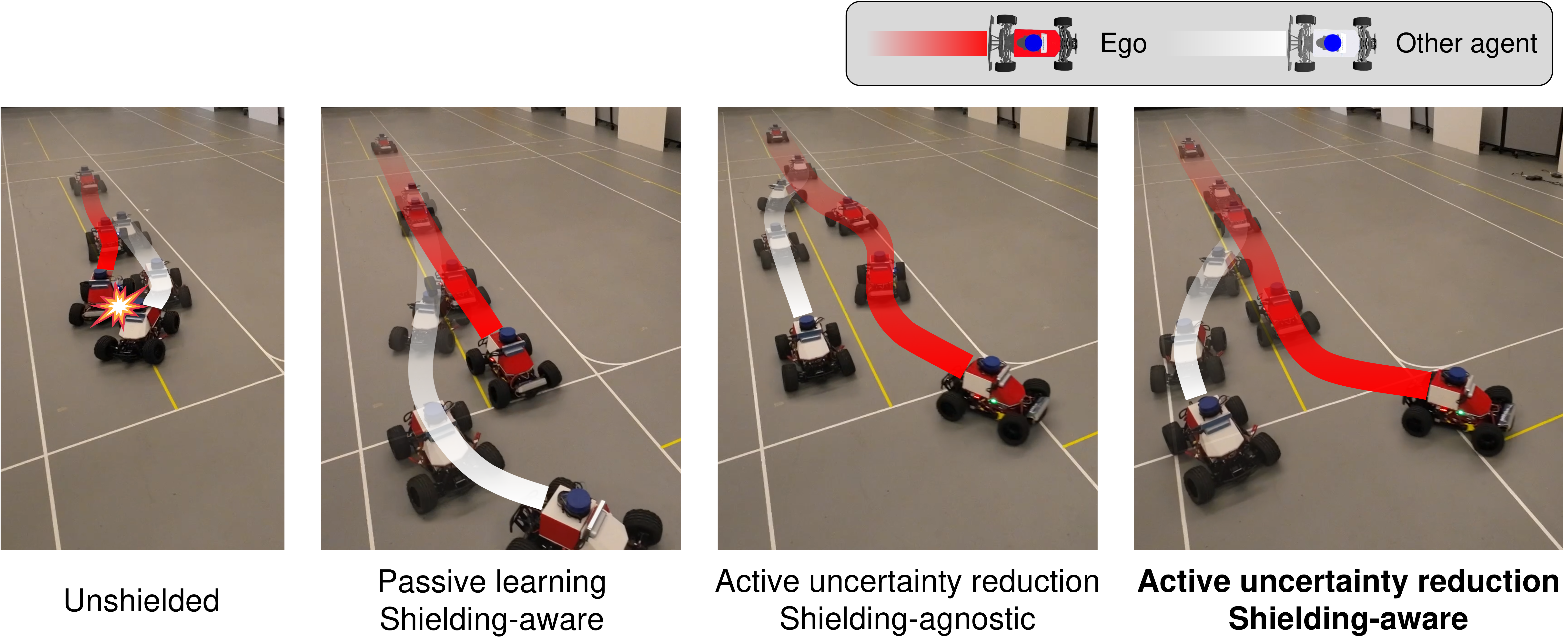}
  \caption{\label{fig:head} Safety and uncertainty reduction are two central components of interaction planning tasks, such as autonomous driving.
  Here, the ego autonomous car (red-front) seeks to overtake the other agent (white-front) autonomous car, whose control policy is unknown to the ego.
  The other car in this case is programmed as a cooperative agent, who will make way for the ego if they are sufficiently close.
  Four trials were controlled with roughly the same initial separation distance between the ego and the other agent.
  \emph{Left}: An unshielded planner without formal safety guarantees is not able to guarantee collision-free maneuvers.
  \emph{Middle-left}: A shielding-aware planner reasons about future shielding events and avoids relying on safety override if possible. However, a lack of active uncertainty reduction can cause the ego agent to be overly conservative and fail to complete the overtaking task. 
  \emph{Middle-right}: A shielding-agnostic policy with active uncertainty reduction produces aggressive overtaking maneuvers and triggers ``near-miss'' emergency overrides at the cost of performance and comfort.
  \emph{Right}: Our proposed control framework combines active uncertainty reduction and shielding-aware planning, which balances safety and efficiency, leading to significant performance improvement for the closed-loop system.
  }
\end{figure*}

Multi-stage trajectory optimization with closed-loop Bayesian inference can be generally cast as a stochastic optimal control problem.
An important aspect of stochastic control with hidden states is whether the computed policy generates the so-called \emph{dual control} effect~\cite{feldbaum1960dual,bar1974dual,mesbah2018stochastic}; that is, in the context of interaction planning, whether the robot \emph{actively} seeks to reduce the uncertainty about the other agent's hidden states.
Solution methods for dual stochastic optimal control problems can be categorized into \emph{explicit} approaches~\cite{heirung2015mpc,sadigh2018planning,tian2021anytime}, which reformulate the problem with some form of heuristic probing, and implicit approaches~\cite{bar1974dual,klenske2016dual,arcari2020dual}, which directly tackle the control problem with stochastic dynamic programming.
While explicit dual control problems are in general easier to formulate and solve than their implicit counterparts, designing the probing term and tuning its weighting factor can be non-trivial and may lead to inconsistent performance.
For a comprehensive review of dual control methods, we refer to~\cite{mesbah2018stochastic}.

\noindent \textbf{Contribution:} In this paper, we formulate a broad class of interactive planning problems in the framework of stochastic optimal control and present an approximate solution method using implicit dual stochastic model predictive control (SMPC).
The resulting policy automatically trades off the cost of exploration and exploitation, allowing the robot to actively reduce the uncertainty about the other agent's hidden states without sacrificing expected planning performance.
Our proposed SMPC problem supports both continuous and categorical uncertainty and can be solved using off-the-shelf real-time nonlinear optimization solvers.
To the best of our knowledge, this is the first interactive motion planning framework that performs active uncertainty reduction without requiring an explicit information-gathering strategy or objective.

In order to provide formal safety guarantees for the robotic system and other agents, we use the dual control policy in conjunction with shielding~\cite{hsu2023sf}, a supervisory safety filter scheme that overrides the dual controller with a safe backup policy.
We improve the recently proposed shielding-aware robust planning (SHARP) framework from~\cite{hu2022sharp}, which generates efficient motions by reasoning about future shielding events triggered by low-probability actions of other agents, to explicitly account for their responses to the ego agent's probing actions.
We propose an easy-to-optimize convex constraint that locally captures the system evolution governed by high-cost safety maneuvers, and integrates it into the implicit dual SMPC formulation.
The central ideas of our method are demonstrated in Fig.~\ref{fig:head}.
\emph{Our key insight is that the shielding-aware dual control policy simultaneously gathers useful information by actively engaging with the other agent, while remaining vigilant about the risk of efficiency loss due to costly shielding overrides.}

A preliminary version of this work~\cite{hu2022active} was presented at the \emph{International Workshop on the Algorithmic Foundations of Robotics (WAFR), 2022}. In this revised and extended paper, we provide the following additional contributions: (i) an extension to general interaction planning problems (a superclass of the human-robot interaction planning problem considered in the original paper), (ii) more in-depth explanation, derivation, and analysis of the proposed framework, including more detailed discussions on the choice of other agent's behavior models and properties of the proposed policy, (iii) incorporation of the shielding-aware robust planning scheme, which provides robust safety guarantees and reconciles conflicts between safety overrides and dual control probing actions, (iv) additional simulation results using the Waymo Open Motion Dataset~\cite{sun2020scalability}, and (v) hardware demonstration on 1/10 scale autonomous vehicles.

\begin{table*}[h]
  \centering
  \caption{\label{tab:comparison} Comparison of selected interaction planning methods that explicitly model coupled motions between agents.}
  \label{tab2}
  \begin{tabular}{p{6.3cm}
  >{\centering}p{1.8cm}
  >{\centering}p{1.3cm}
  >{\centering}p{1.3cm}
  >{\centering}p{1.0cm}
  >{\centering}p{2.0cm}
  c
  }
    \toprule
    Features & \cite{schildbach2015scenario} & \cite{sadigh2018planning} & \cite{peters2020inference} & \cite{tian2021anytime} & \cite{sunberg2022improving}  & \textbf{Ours}\\
    \midrule
    Active uncertainty reduction        & \no & \yes & \no & \yes & \yes & \yes\\
    Automatic exploration-exploitation trade-off     & N/A & \no & N/A & \yes & \yes & \yes\\
    Safety guarantees                   & \yes & \no & \no & \no & \no & \yes\\
    Continuous state space              & \yes& \yes & \yes & \no & \yes & \yes\\
    Continuous action space             & \yes & \yes & \yes & \no & \no & \yes\\
    Scales to more than one opponents   & \yes & \no & \yes & \no & \yes & \yes\\
    Can compute policy fully online       & \yes & \yes & \yes & \no & \yes & \yes\\
    Game-theoretic                      & \no & \yes & \yes & \yes & \no & \yes\\
    \bottomrule
  \end{tabular}
\end{table*}

\begin{figure}[!hbtp]
  \centering
  \includegraphics[width=1.0\columnwidth]{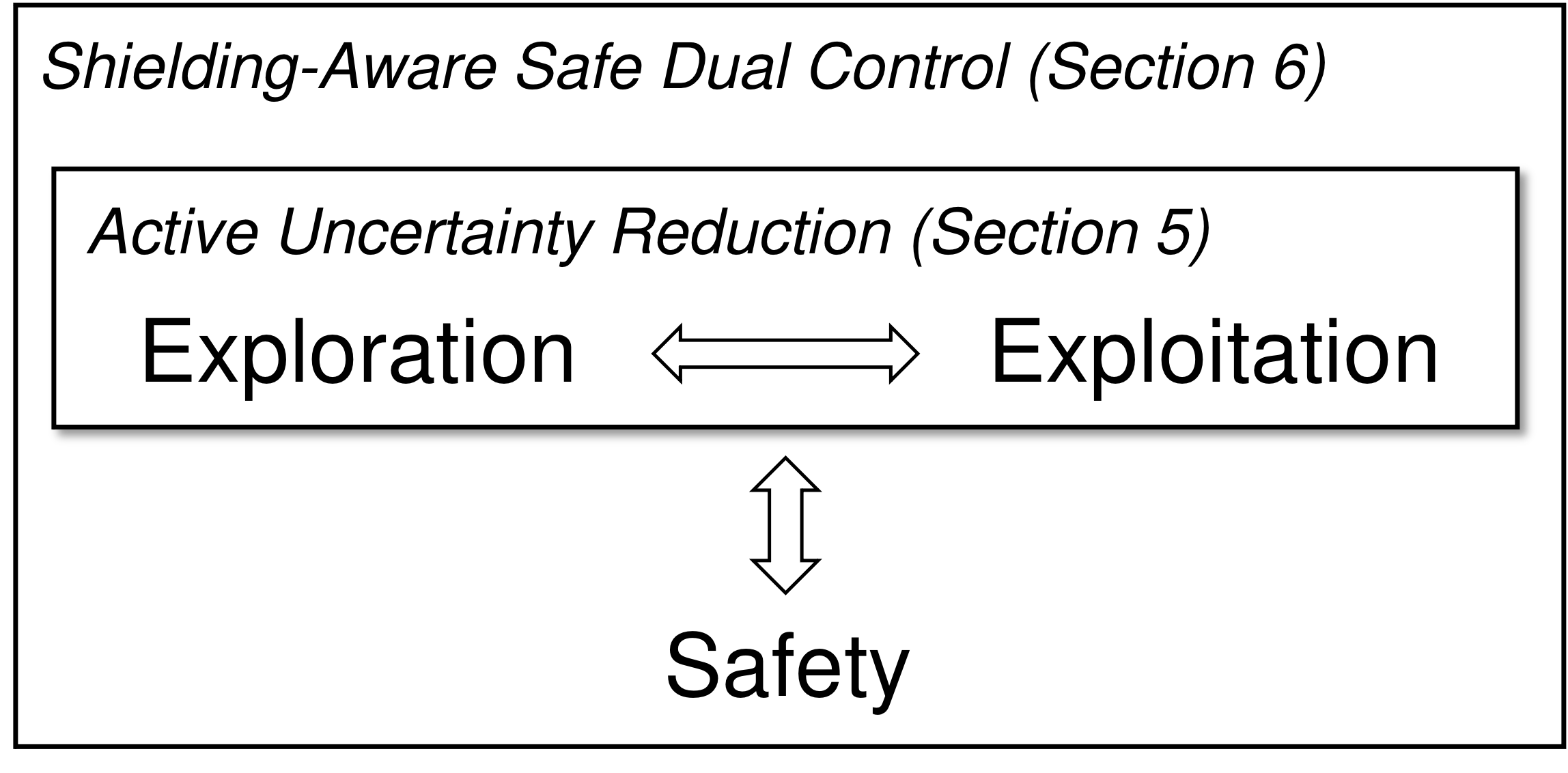}
  \caption{\label{fig:framework} Elements of interaction planning tasks and their relationships studied in the paper.
  In Section~\ref{sec:main}, we focus on dual control--based active uncertainty reduction that improves planning efficiency by balancing exploration and exploitation.
  In Section~\ref{sec:SHARP}, we introduce shielding-aware safe dual control, which reconciles potential conflicts between safety and efficiency. 
  }
\end{figure}

\section{Related Work}
\label{sec:related_work}

In this section, we provide a literature review on different aspects of interaction planning methods.
In Table~\ref{tab:comparison}, we compare planning strategies that explicitly account for the interaction between the ego robot and other agents.
A diagram summarizing our approach is shown in Fig.~\ref{fig:framework}.

\subsection{Dual Control and POMDP}
Robotic motion planning problems that involve identification of other agent's behaviors governed by their unknown intentions can be modeled as a Mixed Observability Markov Decision Process~\cite{bandyopadhyay2013intention}, a variant of the well-known Partially Observable Markov Decision Process (POMDP)~\cite{ong2009pomdps}.
Historically, dual control and POMDP have been studied with different terminologies and application domains by the control theory and machine learning communities, respectively, but later identified as sharing the same fundamental ideas~\cite{dayan1996exploration, bhambri2022reinforcement}.
Due to the intrinsic connection between dual control and POMDP, one problem can usually be cast as the other.
In~\cite{sehr2017tractable}, a dual control problem is reformulated as a POMDP and solved with dynamic programming, which, however, only applies to low-dimensional problems.
On the other hand, computationally more efficient algorithms~\cite{silver2010monte,somani2013despot,sunberg2017value} that rely on Monte Carlo tree search (MCTS) have been developed to approximately solve POMDPs.
In general, solving a dual control problem with MCTS-based POMDP solution methods requires sampled (hence discretized) states, controls, and observations.
Recent work~\cite{sunberg2022improving}, which builds on an earlier algorithm~\cite{sunberg2018online}, extends classical MCTS-based POMDP solution methods to account for continuous state and observation space.
However, dealing with a continuous action space remains an open problem with this approach, and its theoretical information-gathering properties remain unclear~\cite{sunberg2018online}.
In this paper, we instead use trajectory optimization to solve the dual control problem, which allows for directly optimizing states and controls in their \emph{continuous} spaces and comes with \emph{provable} information-gathering properties of the resulting control policy.
When interactions between the robot and the human are explicitly modeled, the POMDP formulation becomes a (usually intractable) Partially-Observable Stochastic Game (POSG)~\cite{sadigh2018planning}.
Our approach can be viewed as a new computationally efficient framework for solving interaction planning problems cast as POSGs.

\subsection{Human-in-the-Loop Motion Planning}
Human-robot interaction, or human-in-the-loop planning, is an emerging and rapidly growing subfield of interaction planning.
Traditional human-robot interaction planning algorithms usually adopt a two-stage pipeline, where humans' future motion is first reasoned with a predictive model, which is then fed into a motion planner.
In~\cite{fisacBHFWTD18}, the human's future trajectories are predicted using the Boltzmann rationality model, and model confidence is concurrently estimated using Bayesian inference.
This information is then used by a motion planner to obtain probabilistic collision-free guarantees. 
This framework is then extended in~\cite{tian2021safety} to explicitly account for game-theoretic models, in which the human's role is also inferred, leading to a less conservative planning performance.
Despite being computationally efficient, the above two-stage pipeline decouples inference from planning, which may lead to conservativeness in planning.
Recently, efforts have been made towards a more integrated framework by solving a joint prediction-planning problem.
In~\cite{sadigh2018planning}, the authors propose to model human-robot interaction as a dynamical system,
in which the robot's action can affect the human's action, thus producing predictions of the human's future states.
Related work in the following years proposes to solve the joint problem as a general-sum dynamic game~\cite{fisac2019hierarchical,fridovich2020efficient,zanardi2021urban}.
Recent work~\cite{hu2023belgame} also considers the joint prediction-planning setting, where the authors focus on safety analysis in belief space, leading to a belief-space zero-sum game formulated using Hamilton-Jacobi (HJ) Reachability.
Our method falls into the category of joint prediction and planning.

\subsection{Dual Stochastic Model Predictive Control} 
SMPC has been widely used for robotic motion planning under uncertainty due to its ability to handle safety-critical constraints and general uncertainty models.
Seminal work~\cite{bernardini2011stabilizing} establishes a general framework for solving SMPC problems with scenario optimization.
In~\cite{schildbach2015scenario}, an SMPC approach is proposed for lane change assistance of the ego vehicle in the presence of other human-driven vehicles, whose future movements are predicted and incorporated as finitely many scenarios in the MPC formulation.
SMPC is also used for autonomous driving at traffic intersections~\cite{nair2021stochastic}, where the trajectories of other vehicles are predicted with Gaussian Mixture Models.
In~\cite{chen2021interactive}, a scenario-based SMPC algorithm is proposed to capture multimodal reactive behaviors of uncontrolled human agents.
In~\cite{hu2022sharp}, a provably safe SMPC planner is developed for robust human-aware robotic motion planning, which proactively balances expected performance with the risk of high-cost emergency safety maneuvers triggered by low-probability human behaviors.
However, all those SMPC methods do not produce dual control effect - the robot only passively adapts to the predicted motion of the other agent but will not actively probe them to gather more information and reduce their uncertainty.
In~\cite{arcari2020dual}, an implicit dual SMPC is proposed for optimal control of nonlinear dynamical systems with both parametric and structural uncertainty.
In~\cite{bonzanini2020safe}, a robust scenario-based SMPC is formulated to enable safe learning of a nonlinear dynamical system, whose dynamics are partially unknown and learned from data based on a Gaussian Process model.
A novel state- and input-dependent scenario tree is used to account for the dependency of the uncertainty model on decision variables, which leads to active uncertainty reduction and improved closed-loop performance.
Our paper builds on those dual SMPC approaches to enable active uncertainty reduction for interaction planning.

\subsection{Active Information Gathering} 
To date, most interaction planning methods follow a ``passively adaptive'' paradigm.
In~\cite{peters2020inference}, a multi-agent interaction planning problem is modeled as a general-sum differential game with equilibrium uncertainty.
The robot first infers which equilibrium the other agent is operating at and then aligns its own strategy with the inferred equilibrium solution.
Recently, the notion of active information gathering, which is conceptually very similar to dual control, has received attention from the robotics, and in particular the human-robot interaction community.
In~\cite{sadigh2018planning}, an additional information gathering term is added to the robot's nominal objective in a trajectory optimization framework for online estimating unknown parameters of the human's state-action value function.
In fact, according to the categorization proposed by~\cite{mesbah2018stochastic}, the method in~\cite{sadigh2018planning} can be classified as an explicit dual control approach, which requires heuristic design of a probing mechanism and weighing the relative importance between optimizing the expected performance and reducing the uncertainty of human's unknown parameters.
On the contrary, we propose an implicit dual control approach, which automatically balances performance with uncertainty reduction, thus not requiring design of a heuristic information gathering mechanism.

\subsection{Safe Interaction Planning via Shielding}
One popular way of improving safety for motion planning under uncertainty is through chance constraint or probabilistically safe planning~\cite{schildbach2015scenario,fisacBHFWTD18,bastani2021statMPS}.
However, using a probabilistically safe planning policy, safety can still be breached when the other agent takes low-probability actions.
This is also known as the issue of the ``long tail'' of unlikely events~\cite{koopman2018heavy}.
All-time safety in interaction planning can be guaranteed by a runtime safety filter, often referred to as \emph{shielding}~\cite{hsu2023sf}.
In this paradigm, a reactive safety fallback policy is used as the ``last resort'', which overrides the nominal policy \textit{only when} a safety-critical event, e.g. a collision, is imminent.
Commonly used shielding mechanisms include, for example, HJ Reachability analysis~\cite{mitchell2005time,bansal2017hamilton,chen2021fastrack}, control barrier functions~\cite{ames2016control,robey2020learning,lindemann2021learning}, model predictive control~\cite{li2020safe,wabersich2021predictive}, and Lyapunov methods~\cite{chow2018lyapunov}.
In~\cite{li2020safe}, robust MPC is used as the backup policy that gives high-probability safety guarantees for a reinforcement learning process.
A unified view of shielding mechanisms can be found in recent survey~\cite{hsu2023sf}.
Despite being effective at guaranteeing safety, shielding can sometimes degrade the planning efficiency, since the safety controllers are typically designed without performance considerations such as task completion time, smoothness of the trajectory or power consumption.
To mitigate this issue, the shielding-aware robust planning (SHARP) framework is developed in~\cite{hu2022sharp}, which proactively balances the nominal planning performance with costly emergency maneuvers triggered by low-probability behaviors of the other agent.
In this paper, we incorporate SHARP into implicit dual SMPC for safe and efficient interaction planning, and improve it by lifting the overly conservative assumption that the other agent ignores the ego agent by explicitly accounting for their responses to the ego agent's probing action.

\section{Preliminaries}
\label{sec:prelim}

\looseness=-1
\subsection{Multi-Agent Dynamical System}
We consider a class of discrete-time input-affine dynamics that capture the interaction between an \emph{ego} robotic system~($\ego$) and one or multiple \emph{other} agents ($\oppo$), e.g., nearby humans,
\begin{equation}
\label{eq:joint_sys}
\state_{t+1} = \dyn (\state_t) + B^\ego(\state_t) \ctrl^\ego_t + B^\oppo(\state_t) \ctrl^\oppo_t + \dstb_t,
\end{equation}
where $\state_t = (\state^\ego_t, \state^\oppo_t) \in \reals^n$ is the joint state vector, $\ctrl^\ego_t \in \cset^\ego \subseteq \reals^{m_\ego}$ and $\ctrl^\oppo_t \in \cset^\oppo \subseteq \reals^{m_\oppo}$ are the control vectors of the ego and other agents, respectively, $f: \reals^{n} \rightarrow \reals^{n}$ is a nonlinear function that describes the autonomous part of the dynamics, $B^\ego: \reals^n \rightarrow \reals^{n \times m_\ego}$ and $B^\oppo: \reals^n \rightarrow \reals^{n \times m_\oppo}$ are control input matrices that can depend on the state, and $\dstb_t$ is an additive uncertainty term representing external disturbance inputs (e.g. wind) and modeling error, which we assume to be Gaussian-distributed: $\dstb_t \sim \gaussian(0, \covar^\dstb)$, i.i.d.

\begin{remark}
Dynamics \eqref{eq:joint_sys} model multiple other agents $\oppo_1, \oppo_2, \ldots$ by concatenating their state and control vectors, i.e. $\state^\oppo = (\state^{\oppo_1}, \state^{\oppo_2}, \ldots)$ and $\ctrl^\oppo = (\ctrl^{\oppo_1}, \ctrl^{\oppo_2}, \ldots)$.
\end{remark}

\subsection{Modeling Agent Behavior}
\label{sec:prelim:oppo_behaviour}
In this paper, we parametrize the other agent's action at each time $t$ as a stochastic policy:
\begin{equation}
\label{eq:exo-agent_ctrl_model}
\textstyle\ctrl^\oppo_t := \sum_{i=1}^{n_\theta} \theta_i^M  \ctrl_i^{M,\oppo}, \quad \subjectto \ctrl^\oppo_t \in \cset^\oppo,
\end{equation}
which is a linear combination of \emph{stochastic basis policies} $\ctrl^{M,\oppo}_i$ with parameter $\theta^M := (\theta^M_1,\theta^M_2,\ldots,\theta^M_{n_\theta}) \in \reals^{n_\theta}$.
We further allow each basis policy $\ctrl^{M,\oppo}_i$ to have different \emph{modes} $M$ that take values from a \emph{finite} set $\mset$, representing different categorical behaviors of the other agent (such as being cooperative, non-cooperative, or unaware). 
We define each basis policy $\ctrl^{M,\oppo}_i$ with the ``noisily-rational'' Boltzmann model from cognitive science~\cite{luce1959individual}.
Under this model, the other agent picks each basis policy according to a probability distribution:
\begin{equation*}
    \ctrl^{M,\oppo}_i \sim 
    \distr\left(\ctrl^{M,\oppo}_i \mid \state, \ctrl^\ego; M\right)=
    \frac{e^{\qfunc_{i}^{M}\left(\ctrl^{M,\oppo}_i; \state, \ctrl^\ego, \right)}}{\int_{\tilde{\ctrl}^\oppo \in \cset^\oppo} e^{\qfunc_{i}^{M}\left(\tilde{\ctrl}^\oppo; \state, \ctrl^\ego\right)} d\tilde{\ctrl}^\oppo },
\end{equation*}
for all $i=1,2,\ldots,n_\theta$ and $M \in \mset$.
Here, $\qfunc_{i}^M(\ctrl^{M,\oppo}_i; \state, \ctrl^\ego)$ is the other agent's basis state-action value (or Q-value) function associated with the $i$-th basis policy and mode $M$.
This model assumes that, for a pair of fixed $(i, M)$, the other agent is exponentially likelier to pick an action that maximizes the state-action value function.

\begin{remark}
Rational decision makers (e.g. robots) are a special case of noisily-rational agents when they choose actions according to $\ctrl^{M,\oppo}_i = \argmax \qfunc_i^M(\cdot)$, i.e. there is no uncertainty (ambiguity) when picking basis policies $\ctrl^{M,\oppo}_i$, while the uncertainties in model parameters $(\theta^M, M)$ and external disturbance inputs $d$ still persist.
Therefore, our framework can naturally account for rational agents.
\end{remark}

\begin{remark}
Our approach is agnostic to the concrete methods for determining the other agent's state-action value function $\qfunc^M_i(\cdot)$, parameters $\theta^M$ and $M$, which are usually specified by the system designer based on domain knowledge or learned from prior data.
Goal-driven models for motion prediction are well-established in the literature. See for example~\cite{sadigh2018planning,ziebart2008maximum}.
We provide two examples below. 
\end{remark}

\begin{figure*}[!hbtp]
  \centering
  \includegraphics[width=2.0\columnwidth]{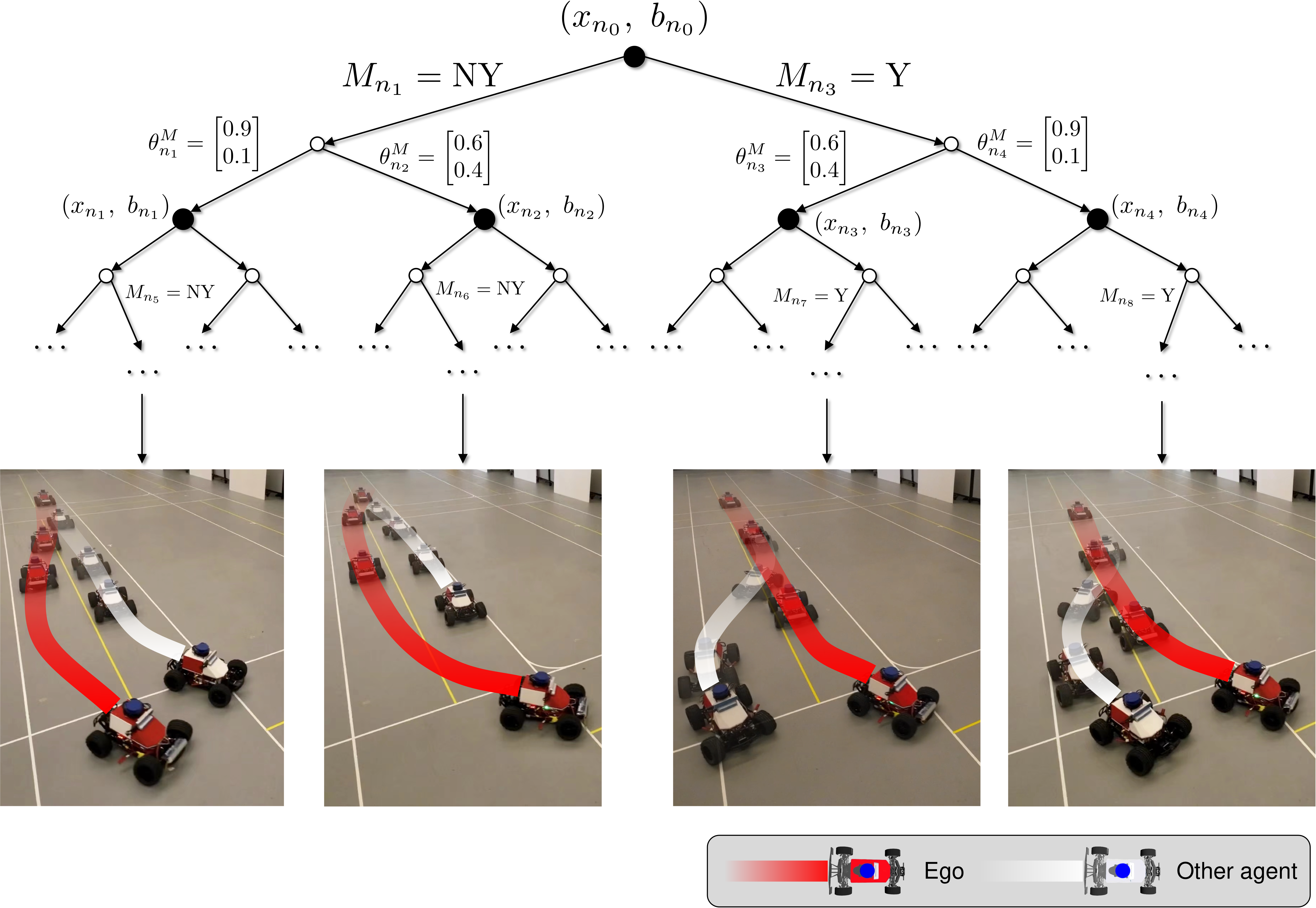}
  \caption{\label{fig:stree}
  Illustration of an optimized scenario tree for Example~\ref{example:1}.
  The ego and other vehicles are painted red front and white front, respectively.
  Uncertainty samples $M_\tnode$ and $\theta^M_\tnode$ associated with node $\tnode$ (see Section~\ref{sec:main:dyn_stree} for scenario tree notations) are used for approximately computing the stochastic objective~\eqref{eq:ST-SMPC:cost},
  and propagating belief states along each scenario branch.
  A white circle $\circ$ denotes an intermediate node partially determined by an $M_\tnode$ sample, and a larger black circle $\CIRCLE$ denotes a node fully determined by both $M_\tnode$ and $\theta^M_\tnode$ samples.
  \emph{Left scenario:} The other agent did not yield, and was less inclined to avoid colliding with the ego vehicle (sampled continuous hidden state $\theta^M_{\text{tr}} \gg \theta^M_{\text{sa}}$). 
  \emph{Middle-left scenario:} The other agent did not yield, but slowed down to avoid the ego vehicle.
  \emph{Middle-right scenario:} The other agent yielded and made a wider turn to avoid hitting the ego vehicle.
  \emph{Right scenario:} The other agent yielded, but was less inclined to avoid the ego vehicle.
  \vspace{-3mm}
  }
\end{figure*}

\begin{example}[\textbf{(Autonomous agent with an unknown policy)}]
\label{example:1}
Consider on an RC car test track (Figure~\ref{fig:head}~and~\ref{fig:hardware}) the ego car is tasked to overtake a robot car controlled by an intent-driven and optimized-based policy that is unknown to the ego.
Following Section~\ref{sec:prelim:oppo_behaviour}, we parametrize the other agent's state-action value function as
\begin{equation*}
    u^\oppo_t = \theta^M_{\rm{tr}} \ctrl^M_{\rm{tr}} \left(\state_t, u^\ego_t\right) + \theta^M_{\rm{sa}} \ctrl^M_{\rm{sa}} \left(\state_t, u^\ego_t\right),~~\text{s.t. } \ctrl^\oppo_t \in \cset^\oppo,
\end{equation*}
where $\theta^M := (\theta^M_{\rm{tr}}, \theta^M_{\rm{sa}})$,
basis policies $\ctrl^M_{\rm{tr}}(\cdot)$ and $\ctrl^M_{\rm{sa}}(\cdot)$ capture the other agent's tracking (e.g. following the reference trajectory) and safety (e.g. avoiding collision with the ego) objectives, respectively.
Modeling the other agent's level of commitment to safety as a \emph{continuum} is motivated by recent work~\cite{toghi2021cooperative, toghi2021social}.
Further, the other agent has two distinct modes, namely willing to yield to the ego by changing the lane or not, i.e. $M \in \{\rm{Y}, \rm{NY}\}$.
An illustration of the other agent's behaviors modeled by different modes and configurations of basis policies can be found in Fig.~\ref{fig:stree}.
\end{example}

\begin{example}[\textbf{(Game-theoretic human-robot interaction)}]
\label{example:2}
In the second example, we consider a traffic intersection scenario that involves an autonomous vehicle (ego) and a pedestrian (the other agent).
We model the pedestrian as a game-theoretic decision maker using model~\eqref{eq:exo-agent_ctrl_model} with parameter $\theta^M := (\theta^M_{\rm{C}}, \theta^M_{\rm{NC}})$, which captures the other agent's level of cooperativeness.
A cooperative agent also optimizes for the robot's objective while a non-cooperative agent does not.
We define the discrete modes $M$ as different (game-theoretic) interactive behaviors of the other agent toward the robot, similar to~\cite{bandyopadhyay2013intention,tian2021safety}.
Specifically, the other agent's state-action value function is defined as
\begin{equation*}
    \qfunc^M_i= \left\{\begin{array}{l}
    \begin{aligned}
    &\qfunc^M_i\left(\ctrl^{M,\oppo}_i; \state_t, \ctrl^{\ego,\text{Nash}}_t(\state_t)\right), \ && \text{if } M = \textnormal{N},\\
    &\qfunc^M_i\left(\ctrl^{M,\oppo}_i; \state_t, \ctrl^{\ego,\text{worst}}_t(\state_t)\right), \ && \text{if } M = \textnormal{p},\\
    &\qfunc^M_i\left(\ctrl^{M,\oppo}_i; \state_t, \ctrl^{\ego,\text{best}}_t(\state_t)\right), \ && \text{if } M = \textnormal{w},\\
    &\qfunc^M_i\left(\ctrl^{M,\oppo}_i; \state^\oppo_t\right), \ && \text{if } M = \textnormal{o}.
    \end{aligned}
    \end{array}\right.
\end{equation*}
In the first three modes, the other agent assumes that the robot's control $\ctrl^\ego_t$ is a (local) feedback Nash equilibrium solution~\cite{bacsar1998dynamic} ($M = \rm{N}$), the worst-case one that minimizes $\qfunc^M_i(\cdot)$ (i.e. a protected agent with $M = \rm{p}$), and the best-case one that maximizes $\qfunc^M_i(\cdot)$ (i.e. a wishful agent with $M = \rm{w}$), respectively; the last mode follows the same assumption as in~\cite{fisacBHFWTD18,bajcsy2021analyzing,hu2022sharp} that the other agent is oblivious and ignores the presence of the robot ($M = \rm{o}$).
\end{example}

\subsection{Inferring Model Parameter}
In general, parameter $\theta^M$ and mode $M$ in action model \eqref{eq:exo-agent_ctrl_model} are \emph{hidden states} that are unknown to the robot.
Therefore, they can only be inferred from past observations.
To address this, we define the information vector $\ivec_t := \left[\state_{t}, \ctrl^\ego_{t-1}, \ivec_{t-1}\right]$ as the collection of all information that is \textit{causally observable} by the robot at time $t \geq 0$, with $\ivec_{0}=\left[\state_{0}\right]$.
We then define the \textit{belief state} ${\bel_t := \distr\left(\theta^M, M \mid \ivec_{t}\right)}$ as the joint distribution of $(\theta^M,M)$ conditioned on $\ivec_t$, and $\bel_0:=\distr\left(\theta^M, M \right)$ is a given prior distribution.
When the ego agent receives a new observation $\state_{t+1}\in \ivec_{t+1}$, the current belief state $\bel_t$ is updated using the recursive Bayesian inference equations:
\begin{subequations}
\begin{align}
&\distr(\theta^M_{-} \mid \ivec_{t+1}; M) \notag \\
&\quad = \frac{ \distr(\state_{t+1} \mid \ctrl^\ego_t, \ivec_t; \theta^M, M) \distr(\theta^M \mid \ivec_{t}; M) }{ \distr(\state_{t+1} \mid \ctrl^\ego_t, \ivec_t; M)} \label{eq:Bayes_est_meas_theta}, \\
&\distr(M_{-} \mid \ivec_{t+1}) = \frac{\distr(\state_{t+1} \mid \ctrl^\ego_t, \ivec_t; M) \distr(M \mid \ivec_{t}) }{ \distr(\state_{t+1} \mid \ctrl^\ego_t, \ivec_t)} \label{eq:Bayes_est_meas_M}, \\
&\bel^-_{t+1} := \distr(\theta^{M}_{-} \mid \ivec_{t+1}; M) \distr(M_{-} \mid \ivec_{t+1}), \label{eq:Bayes_est_param} \\
&\bel_{t+1} = g^t(\bel^-_{t+1}) := \textstyle \int  \distr(\theta^M, M \mid \tilde{\theta}^M_{-}, \tilde{M}_{-}) \notag \\
&\qquad \qquad \qquad \qquad \; \cdot \distr(\tilde{\theta}^M_{-}, \tilde{M}_{-} \mid \ivec_{t+1}) d \tilde{\theta}^M_{-} d \tilde{M}_{-}. \label{eq:Bayes_est_time} 
\end{align}
\end{subequations}
where $\bel^-_{t+1}$ is the belief state updated with the likelihood $\distr(\state_{t+1} \mid \ctrl^\ego_t, \ivec_t; M)$, $\distr(\theta^M, M \mid \tilde{\theta}^M_{-}, \tilde{M}_{-})$ is a transition model and $g^t(\cdot)$ is the belief state transition dynamics.
We can compactly rewrite~\eqref{eq:Bayes_est_meas_theta}-\eqref{eq:Bayes_est_time} as a dynamical system,
\begin{equation}
\label{eq:belief_state_dyn}
{\bel}_{t+1}=g({\bel}_{t}, \state_{t+1}, \ctrl^\ego_{t} ).
\end{equation}
Unfortunately, system \eqref{eq:belief_state_dyn} in general does not adopt an analytical form beyond one-step evolution.
Even if the prior distribution $\bel_t$ is a Gaussian, the posterior $\bel_{t+1}$ ceases to be a Gaussian since the other agent's action $\ctrl^\oppo_t$ defined by \eqref{eq:exo-agent_ctrl_model} (which in turn affects the observation $\state_{t+1}$) is generally non-Gaussian, thus precluding the use of the conjugate-prior properties of Gaussian distributions.
In Section~\ref{sec:main}, we will introduce a computationally efficient method to propagate the belief state dynamics approximately.

\section{Problem Statement}

\subsection{Canonical Interaction Planning Problem}
We now define the central problem we want to solve in this paper: the canonical interaction planning problem, which is formulated as a stochastic finite-horizon optimal control problem as follows:
\begin{subequations}
\label{eq:HRI}
\begin{align}
\label{eq:HRI:obj} \min_{\substack{\Pi^\ego}} \
&\expectation\limits_{\substack{ (\theta^M, M) \sim \bel_{[0:N-1]}, \\ \ctrl_{[0: N-1]}^{\oppo}\sim \eqref{eq:exo-agent_ctrl_model}, \dstb_{[0:N-1]} } } 
\sum_{k=0}^{N-1} \ell \left(\state_k, \policy^\ego_k(\state_k,\bel_k)\right) + \ell_F (\state_N) \\
\text{s.t.} \quad \label{eq:HRI:sys_init} &\state_{0}=\hat{\state}_{t}, \ \bel_{0}=\hat{\bel}_{t}, \\
& \forall k = 0,\ldots,N-1: \notag \\
\label{eq:HRI:sys_dyn} &\state_{k+1} = \dyn (\state_k) + B^\ego \policy^\ego_k + B^\oppo \ctrl^\oppo_k + \dstb_k, &&\hspace{0.0cm}\\
\label{eq:HRI:belief_dyn} &{\bel}_{k+1}=g\left({\bel}_{k}, \state_{k+1}, \policy^\ego_{k}(\state_k, \bel_k)\right),  &&\hspace{0.0cm} \\
\label{eq:HRI:safety} &\state_k \notin \failure, \qquad \forall k = 0,\ldots,N
\end{align}
\end{subequations}
where $\Pi^\ego = \{\policy^\ego_0(\cdot), \ldots, \policy^\ego_{N-1}(\cdot)\}$ is the policy sequence obtained by optimizing \eqref{eq:HRI}, $\hat{\state}_{t}$ and $\hat{\bel}_{t}$ are the state measured and belief state maintained at \emph{real-world time} $t$, the \emph{prediction time} associated with decision variables is indexed with $k$, $\ell: \reals^{n} \times \cset^\ego \rightarrow \reals_{\geq 0}$ and $\ell_F: \reals^{n} \rightarrow \reals_{\geq 0}$ are designer-specified stage and terminal cost function, $\policy_k(\state_k, \bel_k)$ is a \emph{causal} feedback policy~\cite{bar1974dual,mesbah2018stochastic} that leverages the (yet-to-be-acquired) knowledge of future states $\state_{[k,\ldots,N]}$ and belief states $\bel_{[k,\ldots,N-1]}$, and $\failure \subseteq \reals^n$ is a failure set that the state is not allowed to enter.

For the moment, we drop the safety constraint~\eqref{eq:HRI:safety} and defer the discussion of how to tractably enforce it to Section~\ref{sec:SHARP}.
Now, problem \eqref{eq:HRI} can be solved using stochastic dynamic programming~\cite{bellman1966dynamic}.
An optimal robot's value function (minimum cost-to-go) $\valfunc_k(\state_k, \bel_k)$ and control policy $\policy_k^{\ego,*}(\state_k, \bel_k)$ can be obtained backward in time using the Bellman recursion,
\begin{equation}
\label{eq:sto_DP}
\begin{aligned}
&\valfunc_k(\state_k, \bel_k) = \\
&\min_{\substack{
\policy^\ego_k(\state_k,\bel_k)}} \ell(\state_k, \policy^\ego_k) + \expectation\limits_{\substack{ (\theta^M, M) \sim \bel_k, \\ \ctrl_k^{\oppo}\sim \eqref{eq:exo-agent_ctrl_model}, \dstb_k } } \left[\valfunc_{k+1}(\state_{k+1}, \bel_{k+1}) \mid \ivec_k \right] \\
& \quad \;\; \text{s.t.} \ \ \eqref{eq:HRI:sys_dyn}-\eqref{eq:HRI:safety}
\end{aligned}
\end{equation}
with terminal condition $\valfunc_N(\state_N,\bel_N) = \ell_F(\state_N)$.

\subsection{Dual Control Effect}
Value function $\valfunc_t(\state_t, \bel_t)$ obtained by solving \eqref{eq:sto_DP} depends on future belief states $\bel_{t^\prime}\; (t^\prime>t)$,
thus giving the optimal policy the ability to affect \emph{future uncertainty} of the other agent quantified by the belief states.
Therefore, the optimal policy $\ctrl^{\ego,*}_t$ of \eqref{eq:sto_DP} possesses the property of \emph{dual control effect}, defined formally in Definition~\ref{def:DC}.
Due to the principle of optimality~\cite{bellman1966dynamic}, the policy achieves an optimal balance between optimizing the robot's expected performance objective~\eqref{eq:HRI:obj} and actively reducing its uncertainty about the other agent.
In other words, the optimal policy of~\eqref{eq:sto_DP} automatically probes other agents to reduce their uncertainty \emph{only} to the extent that doing so improves the robot's expected closed-loop performance.

\begin{definition}[Dual Control Effect]
\label{def:DC}
A control input has dual control effect if it can affect, with nonzero probability,
\begin{enumerate}[(a)]
    \item \label{def:DC_a} At least one $r$th-order ($r \geq 2$) central moment of a hidden state variable~\cite{feldbaum1960dual,bar1974dual,mesbah2018stochastic}, or
    \item \label{def:DC_b} Entropy of a categorical hidden state~\cite{hijab1984entropy}.
\end{enumerate}
\end{definition}

\subsection{Approximate Dual Control}
Unfortunately, \eqref{eq:sto_DP} is computationally intractable in all but the simplest cases, mainly due to nested optimization of robot's action and computing the conditional expectation.
The expectation term in \eqref{eq:sto_DP} can be approximated to arbitrary accuracy with quantization of the belief states, which, however, leads to exponential growth in computation, i.e. the issue of \emph{curse of dimensionality}~\cite{bellman1966dynamic}.
It is for those reasons that approximate methods are mainly used to solve dual control problems.
Approximate dual control can be categorized into: explicit approaches, e.g.~\cite{heirung2015mpc,sadigh2018planning,tian2021anytime} that simplify the original stochastic optimal control problem by artificially introducing probing effect or information gathering objectives to the control policy, and implicit approaches, e.g.~\cite{bar1974dual,klenske2016dual,arcari2020dual} that rely on direct approximation of the Bellman recursion~\eqref{eq:sto_DP}.
The approach we take in this paper is a \emph{scenario-based implicit dual control method}, which is detailed in the next section.
The main advantage of using the implicit dual control approximation is that the automatic exploration-exploitation trade-off of the policy is naturally preserved in an optimal sense~\cite{sehr2017tractable,mesbah2018stochastic}.

\section{Active Uncertainty Reduction using Implicit Dual Control}
\label{sec:main}
In this section, we describe an implicit dual control approach towards approximately solving the canonical interaction planning problem~\eqref{eq:HRI}.
We start by presenting an approximation scheme for tractably propagating the belief state dynamics and computing the expectation in Bellman recursion~\eqref{eq:sto_DP}.
This is essential for reformulating~\eqref{eq:sto_DP} as a real-time solvable SMPC problem.
The formulation of our proposed SMPC problem and properties of the resulting policy are detailed in Section~\ref{sec:SHARP}.

\subsection{Parameter-Affine Dynamics}
\label{sec:main:param_affine}
In order to tractably propagate belief state dynamics \eqref{eq:belief_state_dyn}, we propose to reformulate joint dynamics~\eqref{eq:joint_sys} into the \emph{parameter-affine} form:
\begin{equation}
\label{eq:param_affine_dyn}
\state_{t+1} = F(\state_t, \ctrl^\ego_t) \theta^M + \bar{\dyn} (\state_t, u^\ego_t) + \bar{\dstb}_t,
\end{equation}
where the control term $B^\oppo(\state_t) \ctrl^\oppo_t$ of the other agent in~\eqref{eq:joint_sys} is replaced with $F(\state_t, \ctrl^\ego_t) \theta^M$, which is linear in parameter $\theta^M$, and $\bar{\dstb}_t$ is a zero-mean Gaussian random variable, which, as we will see, is used to combine Gaussian uncertainty $\dstb_t$ in~\eqref{eq:joint_sys} and the uncertainty in the other agent's action.
This serves as a key building block for deriving the approximate belief updating rule in Section~\ref{sec:main:belief_update}.

To obtain Gaussian parameter-affine dynamics in the form of~\eqref{eq:param_affine_dyn}, the main technical tool we rely on is the Laplace approximation~\cite[Chapter~4]{bishop2006PRML}.
Precisely, the conditional probability distribution of each basis policy $\ctrl^{M,\oppo}_i$ is approximated as:
\begin{equation}
\label{eq:Laplace}
    \distr\left(\ctrl^{M,\oppo}_i \mid \state_t, \ctrl^\ego_t; M\right)
    \approx \gaussian\left(\mu_i^M(\cdot), \covar_i^M(\cdot) \right),
\end{equation}
where the mean function of the basis policy is
\begin{equation*}
    \mu_i^{M}(\state_t, \ctrl^\ego_t) := \textstyle\argmax_{\ctrl_{i}^{M,\oppo} \in \cset^\oppo} \qfunc_{i}^{M}\left(\ctrl_{i}^{M,\oppo}; \state_{t}, \ctrl^\ego_t\right),
\end{equation*}
and the covariance function of the basis policy is
\begin{equation*}
    \covar_i^M(\state_t, \ctrl^\ego_t) := \textstyle -\nabla^2_{\ctrl_i^{M,\oppo}} \left. {\qfunc_{i}^{M}\left(\ctrl_{i}^{M,\oppo}; \state_{t}, \ctrl^\ego_t\right)}^{-1}\right|_{\mu_i^M(\state_t, \ctrl^\ego_t)}.
\end{equation*}
The intuition behind the above Laplace approximation scheme is that the Gaussian distribution obtained in \eqref{eq:Laplace} centers around the \emph{mode} $\mu_i^M(\state_t, \ctrl^\ego_t)$ of the original basis policy distribution $\distr\left(\ctrl_{i}^{M,\oppo} \mid \state_t, \ctrl^\ego_t; M\right)$, which corresponds to the \emph{perfectly rational} action of the other agent associated with $\theta_i^M$.
We discuss maximization of the basis Q-value function $\qfunc_{i}^{M}(\cdot)$ in Appendix~\ref{sec:practical:rational}.
The overall \emph{approximate} action distribution, conditioned on $\theta^M$ and $M$, is given by:
\begin{equation}
\label{eq:exo-agent_ctrl_model_approx}
    \distr\left(\ctrl_t^\oppo \mid \state_t, \ctrl^\ego_t; \theta^M\right)
    \approx \gaussian\left(\mean_t^{\ctrl^\oppo}(\cdot), \covar_t^{\ctrl^\oppo}(\cdot) \right)
\end{equation}
where the mean function of the other agent's policy is
\begin{equation*}
    \mu_t^{\ctrl^\oppo}(\state_t, \ctrl^\ego_t; \theta^M) := \sum_{i=1}^{n_\theta} \theta_i^M \mu_i^M(\state_t, \ctrl^\ego_t),
\end{equation*}
and the covariance function of the other agent's policy is
\begin{equation*}
    \covar_t^{\ctrl^\oppo}(\state_t, \ctrl^\ego_t; \theta^M) := \sum_{i=1}^{n_\theta} \left(\theta_i^M\right)^2 \covar_i^M(\state_t, \ctrl^\ego_t).
\end{equation*}
We subsequently lift the requirement that $\ctrl^\oppo_t \in \cset^\oppo$ in order to keep $\ctrl^\oppo_t$ normally distributed \emph{during belief propagation}, and we use a projected $\ctrl^\oppo_t$ for state evolution.
We discuss how to deal with the unbounded support of predicted $\ctrl^\oppo_t$ in Appendix~\ref{sec:practical:unbounded}.
Based on the covariance function of $\ctrl^\oppo$ obtained above, we can compute the covariance of the disturbance term $\bar{\dstb}_t$ in parameter-affine dynamics~\eqref{eq:param_affine_dyn} as
\begin{equation}
\label{eq:combined_dstb_covar}
\begin{aligned}
\covar^{\bar{\dstb}}_t(\state_t, \ctrl^\ego_t; \theta^M) := \covar^\dstb + B^\oppo \covar_t^{\ctrl^\oppo}(\state_t, \ctrl^\ego_t; \theta^M) {B^\oppo}^\top,
\end{aligned}
\end{equation}
which captures the \emph{combined} uncertainty stemmed from the Gaussian external disturbance $\dstb_t$ in~\eqref{eq:joint_sys} and the other agent's noisily-rational action $\ctrl^\oppo_t$ characterized by~\eqref{eq:exo-agent_ctrl_model_approx}.

As the last step towards obtaining the parameter-affine dynamics, we define the \emph{mean matrix} by concatenating the mean functions of the other agent's basis policies:
\begin{equation}
\label{eq:policy_mat}
    U^\oppo(\state_t, \ctrl^\ego_t) := \begin{bmatrix} \mu_1^M(\state_t, \ctrl^\ego_t) &\ldots &\mu_{n_\theta}^M(\state_t, \ctrl^\ego_t) \end{bmatrix}.
\end{equation}
Plugging~\eqref{eq:exo-agent_ctrl_model} and~\eqref{eq:policy_mat} into~\eqref{eq:joint_sys} leads to:
\begin{equation}
\label{eq:joint_sys_approx}
\begin{aligned}
    \state_{t+1} &= \underbrace{B^\oppo(\state_t) U^\oppo(\state_t, \ctrl^\ego_t)}_{=:F(\state_t, \ctrl^\ego_t)} \theta^M + \underbrace{\dyn (\state_t) + B^\ego(\state_t) \ctrl^\ego_t}_{=:\bar{\dyn} (\state_t, u^\ego_t)} + \bar{\dstb}_t,
\end{aligned}
\end{equation}
which is in form of parameter-affine dynamics~\eqref{eq:param_affine_dyn}.

\begin{remark}
Even if dynamics \eqref{eq:joint_sys_approx} are linear in parameter $\theta^M$, dependence of covariance matrix $\covar^{\bar{\dstb}}_t$ on $\theta^M$, as shown in~\eqref{eq:combined_dstb_covar}, still prohibits updating the belief states in closed-form.
To this end, we approximate $\covar^{\bar{\dstb}}_t$ by fixing $\theta^M$ with some estimated value $\bar{\theta}^M$.
In our paper, we estimate $\bar{\theta}^M$ using a roll-out-based approach by setting its value to the mean of the conditional distribution of $\theta^M$ computed in Step 3 of the initialization pipeline described in Appendix~\ref{sec:practical:init}.
In practice, replacing random variable $\theta^M$ with its point estimate $\bar{\theta}^M$ will inevitably introduce error in covariance $\covar^{\bar{\dstb}}_t$.
\end{remark}

\begin{remark}[Alternative Action Model]
In this paper, we mainly use model~\eqref{eq:exo-agent_ctrl_model} (linear combination of basis policies) for predicting other agents' actions in simulation studies and experiments.
Nonetheless, we highlight that our methodology applies to \emph{any} action model that can lead to parameter-affine dynamics~\eqref{eq:param_affine_dyn}.
Here, we provide another action model that falls into such a category.
Similar to~\cite[Section~IV.A]{bobu2020quantifying}, we parametrize the other agent's state-action value function\footnote{In~\cite{bobu2020quantifying}, the authors used a linear combination of basis cost (i.e. value) functions that only depend on the states instead of state-action value functions to model the other agent's behavior.} as a linear combination of basis functions:
\begin{equation*}
\qfunc_\theta^M\left(\ctrl^\oppo; \state, \ctrl^\ego, \theta^M\right) = \sum_{i=1}^{n_\theta} \theta_i^M \qfunc^M_i\left(\ctrl^\oppo; \state, \ctrl^\ego\right),
\end{equation*}
and an associated noisily-rational Boltzmann model:
\begin{equation*}
\begin{aligned}
    &\textstyle\ctrl^\oppo_t \sim 
    \distr\left(\ctrl^\oppo \mid \state, \ctrl^\ego; \theta^M\right)=
    \frac{e^{\qfunc^{M}_\theta\left(\ctrl^\oppo; \state, \ctrl^\ego, \theta^M\right)}}{\int_{\tilde{\ctrl}^\oppo \in {\cset}^{\oppo}} e^{\qfunc^{M}_\theta\left(\tilde{\ctrl}^\oppo; \state, \ctrl^\ego, \theta^M\right)} d\tilde{\ctrl}^\oppo }.
\end{aligned}
\end{equation*}
We again leverage the Laplace approximation to locally approximate the above model as $\distr\left(\ctrl_t^\oppo \mid \state_t, \ctrl^\ego_t; \theta^M\right) \approx \gaussian\left(\mean_t^{\ctrl^\oppo}(\state_t, \ctrl^\ego_t; \theta^M), \covar_t^{\ctrl^\oppo}(\state_t, \ctrl^\ego_t; \theta^M) \right).$
We then perform a first-order Taylor expansion around a given $\bar{\theta}^M$:
\begin{equation*}
\begin{aligned}
\mean_t^{\ctrl^\oppo} (\cdot) &\approx \left.{\nabla_{\theta^M} \mean_t^{\ctrl^\oppo}\left(\state, \ctrl^\ego; \theta^M\right)}\right|_{\bar{\theta}^M} \delta \theta^M + \bar{\mean}_t^{\ctrl^\oppo}(\state_t, \ctrl^\ego_t; \bar{\theta}^M) \\
&=: U^\oppo (\state, \ctrl^\ego) \delta \theta^M + \bar{\mean}_t^{\ctrl^\oppo}(\state_t, \ctrl^\ego_t; \bar{\theta}^M),
\end{aligned}
\end{equation*}
where $\delta \theta^M := \theta^M - \bar{\theta}^M$ is the error term.
This leads to parameter-affine dynamics in form of \eqref{eq:param_affine_dyn} with $F(\state_t, \ctrl^\ego_t) = B^\oppo(\state_t) U^\oppo (\state_t, \ctrl_t^\ego)$ and $\bar{f}(\state_t, \ctrl^\ego_t) = \dyn(\state_t) + B^\ego(\state_t) \ctrl^\ego_t + B^\oppo(\state_t) \bar{\mean}_t^{\ctrl^\oppo}(\state_t, \ctrl^\ego_t; \bar{\theta}^M) - F(\state_t, \ctrl^\ego_t) \bar{\theta}^M$.
\end{remark}

\subsection{Tractable Reformulation of Belief Updates}
\label{sec:main:belief_update}

Given the parameter-affine dynamics in Section~\ref{sec:main:param_affine}, we can now derive a tractable recursive update rule for the belief state dynamics \eqref{eq:belief_state_dyn}.
Since the support of $\theta^M$ is continuous, we model its conditional belief as a Gaussian, i.e. $\distr(\theta^{M} \mid \ivec_{t}; M) \sim \gaussian(\mean^{\theta^M}_{t}, \covar^{\theta^M}_{t})$.
Likewise, $\distr(M \mid \ivec_{t})$ is modeled as a categorical distribution due to its discrete support $\mset$.

Our central idea is to update $\distr(\theta^{M} \mid \cdot)$ efficiently leveraging the self-conjugate property~\cite[Appendix~B]{bishop2006PRML} of Gaussian distributions, that is, given Gaussian prior $\distr(\theta^{M} \mid \ivec_{t}; M)$, if the likelihood $\distr(\state_{t+1} \mid \ctrl^\ego_t, \ivec_t; \theta^M, M)$ is Gaussian, then the posterior $\distr(\theta^M_{-} \mid \ivec_{t+1}; M)$ is also Gaussian, whose mean and covariance are analytical functions of state $\state_t$ and ego's action $\ctrl^\ego_t$.
The following lemma summarizes the approximate belief update procedure for $\distr(\theta^{M} \mid \cdot)$, whose proof can be found in Appendix~\ref{apdx:proof:theta_update}.

\begin{lemma}
\label{lem:theta_update}
For Gaussian-noisy dynamics~\eqref{eq:param_affine_dyn} and policy distribution $\distr\left(\ctrl_t^\oppo \mid \state_t, \ctrl^\ego_t; \theta^M\right)$ of the other agent in~\eqref{eq:Laplace}, if the prior distribution of $\theta^M$ is given by $\distr(\theta^{M} \mid \ivec_{t}; M) \sim \gaussian(\mean^{\theta^M}_{t}, \covar^{\theta^M}_{t})$, then the posterior  distribution is $\distr(\theta^M_{-} \mid \ivec_{t+1}; M) \sim \gaussian(\mean^{\theta^M_{-}}_{t+1}, \covar^{\theta^M_{-}}_{t+1})$ whose mean and covariance are given by
\begin{equation*}
\begin{aligned}
    \mean^{\theta^M_{-}}_{t+1} &= \covar^{\theta^M_{-}}_{t+1} \left[ {F\left( \state_t, \ctrl^\ego_t \right)}^\top \left(\covar^{\state}_{t+1}\right)^{-1} \left( x_{t+1} - \bar{\dyn} (\state_t, u^\ego_t) \right) \right. \\
    &\hspace{1.5cm} \left. + \left(\covar^{\theta^M}_{t}\right)^{-1} \mean^{\theta^M}_{t} \right]\\
    \covar^{\theta^M_{-}}_{t+1} &= \left[ \left(\covar^{\theta^M}_{t}\right)^{-1} + F\left( \state_t, \ctrl^\ego_t\right)^\top \left(\covar^{\state}_{t+1}\right)^{-1} F\left( \state_t, \ctrl^\ego_t\right)\right]^{-1}
\end{aligned}
\end{equation*}
where $\covar^\state_{t+1} := \covar^\dstb + B^\oppo(\state_t) \covar_t^{\ctrl^\oppo}(\state_t, \ctrl^\ego_t; \theta^M){B^\oppo(\state_t)}^\top$.
\end{lemma}

\noindent Importantly, the approximate belief state dynamics in Lemma~\ref{lem:theta_update} serve as an essential step towards preserving the dual control effect when approximately solving stochastic dynamic programming~\eqref{eq:sto_DP}.
This can be intuitively seen by inspecting that the robot's control $\ctrl^\ego_t$ enters the updating equation of covariance matrix $\covar^{\theta^M_{-}}_{t+1}$.
Therefore, $\ctrl^\ego_t$ affects $\covar^{\theta^M_{-}}_{t+1}$ and all future covariance matrices implicitly by affecting future states $\state_{t^\prime}$\; ($t^\prime>t+1$),
hence producing dual control effect for $\theta^M$ according to Definition~\ref{def:DC}.
A formal proof of dual control effect provided by the SMPC policy using the approximate belief state dynamics in Lemma~\ref{lem:theta_update} can be found in Section~\ref{sec:SHARP} (Theorem~\ref{thm:dual_control}).
Measurement update \eqref{eq:Bayes_est_meas_M} for $M$ can be readily computed by marginalizing the likelihood with respect to $\theta^M$ and subsequently applying the Bayes rule, similar to~\cite{arcari2020dual}.
We denote the approximate belief state dynamics in compact form as
\begin{equation}
\label{eq:appox_bstate_dyn}
    {\bel}_{t+1}=\tilde{g} \left({\bel}_{t}, \state_{t+1}, \ctrl^\ego_{t} \right),
\end{equation}
which will later be used in SMPC as dynamics constraints for computationally tractable belief propagation.

\subsection{Dynamic Scenario Trees}
\label{sec:main:dyn_stree}
In this section, we propose an approximate solution method for the canonical interaction planning problem \eqref{eq:HRI} using scenario tree--based stochastic model predictive control (ST-SMPC)~\cite{bernardini2011stabilizing}, which yields a control policy with dual control effect.
The key idea of ST-SMPC is to approximate the expectation in Bellman recursion \eqref{eq:sto_DP} based on uncertainty samples, i.e. quantized belief states.
This leads to a scenario tree that allows us to roll out~\eqref{eq:sto_DP} as a deterministic finite-horizon optimal control problem, which can be readily solved by gradient-based algorithms.
Since our approach hinges on directly approximating Bellman recursion~\eqref{eq:sto_DP}, it can be understood as an implicit dual control method~\cite{mesbah2018stochastic}.
Unlike conventional ST-SMPC methods in which uncertainty samples are fixed during optimization, such as~\cite{lucia2013multi,schildbach2015scenario,hu2022sharp},
our scenario tree has both \emph{state- and input-dependent} uncertainty realizations, leading to a \emph{dynamic} scenario tree.
As a result, uncertainty samples can adjust their value in response to predicted states and inputs during online optimization.
Fundamentally, it is this feature that allows the ego agent to interact with the other agent in a way that promotes the dual control effect, which is explained in detail in Theorem~\ref{thm:dual_control}.

We denote a \emph{node} in the scenario tree as $\node$, whose time, state, and belief state are denoted as $t_n$, $\state_n$, and $\bel_\node$, respectively.
Similarly, the uncertainty samples of the node are $\theta^M_n$, $\bar{\dstb}^M_n$, and $M_n$.
Here, recall that $\bar{\dstb}^M_n$ is the combined disturbance defined in~\eqref{eq:combined_dstb_covar}, which implicitly characterizes a sample of the other agent's action.
The set of all nodes is defined as $\nodeset$.
We define the transition probability from a parent node $\pre{\node}$ to its child node $\node$ as $\bar{\prob}_{\node} := \prob(\theta^M_n \mid \ivec_{\pre{\node}}; M_\node) \prob(\bar{\dstb}^{M}_n \mid \ivec_{\pre{\node}}; M_\node) \prob\left(M_\node \mid \ivec_{\pre{\node}}\right).$
Subsequently, the \textit{path transition probability} of node $\node$, i.e. the transition probability from the root node $\node_0$ to node $\node$ can be computed recursively as $\prob_{\node} := \bar{\prob}_{\node} \cdot \bar{\prob}_{\pre{\node}} \cdots \bar{\prob}_{\node_0}$, with $\bar{\prob}_{\node_0}=1$.

\begin{algorithm}[!hbtp]
	\caption{Construct a scenario tree}
	\label{alg:tree}
	\begin{algorithmic}[1]
	\Require Current state $\hat{x}_t$ and belief state $\hat{b}_t$, horizon $N > 0$, dual control horizon $1 \leq N^d \leq N$, mode set $\mset$, branching number $K > 0$
	\Ensure A scenario tree defined by node set $\nodeset_t$
	\LineComment{Initialization:}
	\State Set $\state_{n_0} \gets \hat{\state}_t$,~$\bel_{n_0} \gets \hat{b}_t$,~$t_{n_0} \gets 0$,~$\nodeset_t \gets \{n_0\}$
	\LineComment{Dual Control Steps:}
	\ForAll{$t^\prime = 0, 1, \ldots, N^d$}
        \ForAll{$\tnode \gets \nodeset_t$}
	        \If{$\tnode.t = t^\prime$}
	            \State Branching: $\nodeset_t \gets \nodeset_t \cup \textsc{Branch}(\tilde{n},\mset,K)$
            \EndIf
	    \EndFor
    \EndFor
    \LineComment{Exploitation Steps:}
    \ForAll{$t^\prime = N^d, N^d+1, \ldots, N$}
        \ForAll{$\tnode \gets \nodeset_t$}
	        \If{$\tnode.t = t^\prime$}
	            \State Extending: $\nodeset_t \gets \nodeset_t \cup \textsc{Extend}(\tilde{n})$
            \EndIf
	    \EndFor
    \EndFor
	\end{algorithmic}
\end{algorithm}

\begin{algorithm}[!hbtp]
	\caption{Generate child nodes for dual control steps}
	\label{alg:branch}
	\begin{algorithmic}[1]
	\Function \textsc{Branch}$(n, \mset, K)$
	\State Initialize the set of child nodes: $\cnodeset \gets \emptyset$
	\ForAll{$M \gets \mset$}
	    \State Randomly sample a set $\{\theta^{M,o}_1,\theta^{M,o}_2,\ldots,\theta^{M,o}_K\}$ from the standard Gaussian $\gaussian(0,I)$
	    \State Randomly sample a set $\{\bar{\dstb}^{M,o}_1,\bar{\dstb}^{M,o}_2,\ldots,\bar{\dstb}^{M,o}_K\}$ from the standard Gaussian $\gaussian(0,I)$
	    \ForAll{$k \gets 1,2,\ldots,K$}
	        \State Create a node $\tnode$
	        \State $\theta^{M,o}_\tnode = \theta^{M,o}_k$, $\bar{\dstb}^{M,o}_\tnode = \bar{\dstb}^{M,o}_k$, $M_\tnode = M$
	        \State $t_\tnode = t_\node + 1$
	        \State $\cnodeset \gets \cnodeset \cup \{\tnode\}$
	    \EndFor
	\EndFor
	\State \Return $\cnodeset$
	\end{algorithmic}
\end{algorithm}

\begin{algorithm}[!hbtp]
	\caption{Generate a child node for exploitation steps}
	\label{alg:extend}
	\begin{algorithmic}[1]
	\Function \textsc{Extend}$(n)$
	\State Create a node $\tnode$
	\State $t_\tnode = t_\node + 1$, $\theta^{M,o}_\tnode = 0$, $\bar{\dstb}^{M,o}_\tnode = 0$, $M_\tnode = M_\node$
	\State \Return $\{\tnode\}$
	\end{algorithmic}
\end{algorithm}

In order to quickly compute the conditional probabilities of $(\theta^M_n, \bar{\dstb}^{M}_n)$ and avoid online sampling (i.e. during the optimization), we use an offline sampling procedure, leveraging the fact that they are (conditional) Gaussian random variables, similar to what is done in~\cite{arcari2020approximate, bonzanini2020safe}.
We first generate samples offline from the standard Gaussian distribution. \remove{, i.e. $\theta^{M,o}_n \sim \gaussian(0,I)$ and $\bar{\dstb}^{M,o}_n \sim \gaussian(0,I)$.}
Then, during online optimization, these samples are transformed using the analytical mean and covariance expressions:
\begin{subequations}
\begin{align}
    \label{eq:sample_trans_theta}
    \theta^M_n &= \mean^{\theta^M_{\node}}(\state_\node, \ctrl^\ego_\node) + \left(\covar^{\theta^M_{\node}}(\state_\node, \ctrl^\ego_\node)\right)^{1/2} \theta^{M,o}_n, \\
    \label{eq:sample_trans_dstb}
    \bar{\dstb}^{M}_n &=  \left(\tilde{\covar}^{\bar{\dstb}^M_\node} (\state_\node, \ctrl^\ego_\node)\right)^{1/2} \bar{\dstb}^{M,o}_n.
\end{align}
\end{subequations}
This reveals the dynamic nature of our proposed scenario tree: the uncertainty samples are adjustable during optimization via transformations~\eqref{eq:sample_trans_theta} and~\eqref{eq:sample_trans_dstb}; the path transition probabilities are also state- and input-dependent due to the Bayesian update of $M$ in~\eqref{eq:Bayes_est_meas_M}.
The scenario tree construction procedure is summarized in Alg.~\ref{alg:tree}.

\begin{remark}
In order to alleviate the exponential growth of complexity associated with the scenario tree, only a small number of $\theta^{M,o}_n$, $\bar{\dstb}^{M,o}_n$ and $M$ are sampled.
We developed a scenario pruning mechanism in our prior work~\cite{hu2022sharp} for static scenario trees, which can be used here as a heuristic to prune branches based on the last optimized scenario tree.
Principled pruning methods for dynamic scenario trees involving state- and input-dependent uncertainty realizations remain an open question.
\end{remark}

\subsection{Exploitation Steps}
In order to alleviate the computation challenge caused by the exponential growth of nodes in the scenario tree, we can stop branching the tree at a stage $N^d < N$, which we refer to as the dual control horizon. 
Subsequently, the remaining $N^e := N - N^d$ stages become the exploitation horizon, where each scenario is extended without branching
and the belief states are only propagated with the transition dynamics $g^t(\cdot)$ defined in~\eqref{eq:Bayes_est_time}, corresponding to a non-dual SMPC problem.
Thanks to the scenario tree structure, control inputs of the exploitation steps still preserve the causal feedback property, allowing the robot to be cautious and ``passively adaptive'' to future uncertainty realizations of the other agent.

\section{Shielding-Aware Safe Dual Control}
\label{sec:SHARP}
In this section, we revisit safety constraint~\eqref{eq:HRI:safety} and provide safety guarantees for the closed-loop system using the proposed implicit dual SMPC policy.
We outline the main ideas of this section below.
First, to ensure that $\state_t \notin \failure$ for all $t \geq 0$ despite the \textit{worst-case} actions of the other agent and external disturbances\footnote{This can also be interpreted as the maximal model mismatch when using~\eqref{eq:exo-agent_ctrl_model} to predict the other agent's action and future state evolution.}, we use a class of least-restrictive supervisory control schemes, often referred to as \textit{shielding}.
Examples of shielding methods and applications of shielding for safe control include, but not limited to:~\cite{bansal2017hamilton, fisac2018general, bonzanini2020safe, bastani2021safe, wabersich2021predictive, hu2022sharp}.
This approach relies on a safety fallback policy, usually as the ``last resort'', which overrides a nominal policy when a safety-critical event, e.g. a collision, is imminent.
The \emph{reactive} nature of a shielding policy makes it agnostic to ego's planning efficiency, and oftentimes conflicts with the probing actions generated by the dual control policy, leading to performance degradation in planning.
To mitigate the conflict between safety overrides and active uncertainty reduction, we augment the implicit dual SMPC method developed in Section~\ref{sec:main} with the recently proposed shielding-aware robust planning (SHARP) framework~\cite{hu2022sharp}.
The key feature of SHARP is that it makes use of the propagated belief states to causally anticipate possible shielding events in the future, thereby \emph{proactively} balancing the nominal planning performance with costly emergency maneuvers triggered by unlikely other agents' behaviors.
The resulting policy ultimately leads to, instead of a low probability of collision, a low chance of having to apply the costly shielding policy.
Note that the original SHARP framework in~\cite{hu2022sharp} assumes that the other agent is oblivious to the ego agent.
In the following, we will lift this overly conservative assumption and explicitly account for other agents' responses.

\subsection{Shielding for Safe Interaction Planning}
In this paper, we focus on providing robust (worst-case) safety guarantees for interaction planning.
Therefore, we make an operational design domain (ODD)~\cite{lee2020identifying} assumption that the external disturbance $\dstb_t$ is bounded.

\begin{assumption}
\label{assump:dset}
The external disturbance $\dstb_t$ is bounded element-wise, i.e. $\dstb_t \in \dset \subseteq \reals^{n_x}$ and $\bar{\dstb}^i_{\min} \leq \dstb^i_t \leq \bar{\dstb}^i_{\max}$ for all $i=1,\ldots,n_x$.
\end{assumption}

A \emph{shielding mechanism} is defined as a tuple $(\Omega, \policy^s)$, where set $\Omega \subseteq \reals^{n_x}$ is a safe set that is \emph{robust controlled-invariant}~\cite{blanchini1999set} and satisfies $\Omega \cap \failure = \emptyset$, and $\policy^s: \reals^{n_x} \rightarrow \cset^\ego$ is a safe control {policy} that keeps the state of joint system~\eqref{eq:joint_sys} inside $\Omega$ even under the worst-case action of the other agent and external disturbance.

\begin{definition}[Robust controlled-invariant set]
\label{def:RCI_set}
Given joint dynamics~\eqref{eq:joint_sys} with bounded actions $\ctrl_t^\oppo\in\cset^\oppo$ of the other agent and uncertain input $\dstb_t\in\dset$,
a set ${\Omega \subseteq \reals^{n_x}}$ is a robust controlled-invariant set if there exists a control policy ${\policy^s: \reals^{n_x}\!\rightarrow \cset^\ego}$ that keeps $\state_t$ from leaving~$\Omega$:
\begin{equation*}
\begin{aligned}
    &\state_0 \in \Omega \\
    &\quad \Rightarrow \state_t \in \Omega,~\forall t>0,~\forall \ctrl^\oppo_t \in \cset^\oppo,~\forall \dstb_t \in \dset,~\ctrl^\ego_t=\policy^\shield(\state_t).
\end{aligned}
\end{equation*}
\end{definition}
\noindent Let $\state^+$ denote the next state evolved with~\eqref{eq:joint_sys} given $\state, \ctrl^\ego, \tilde{\ctrl}^{\oppo}$, and $\tilde{\dstb}$, we subsequently define the \emph{shielding set}:
\begin{equation*}
\begin{aligned}
\label{eq:shield_set}
    &\shieldset^\ego = \{ (\state, \ctrl^\ego) \in \Omega\times\cset^\ego \mid \exists \tilde{\ctrl}^{\oppo} \in \cset^{\oppo},~\exists \tilde{\dstb} \in \cset^{\oppo}: \\
    &\hspace{1.2cm} \state^{+}\left(\state, \ctrl^\ego, \tilde{\ctrl}^{\oppo}, \tilde{\dstb}\right) \notin \Omega \},
\end{aligned}
\end{equation*}
which contains all state-action pairs that \emph{might} lead to the next state departing the safe set.
Based on shielding mechanism $(\Omega, \policy^s)$, we define a least-restrictive supervisory \emph{safety filter}~\cite{hsu2023sf} as a switching policy:
\begin{equation}
\label{eq:safety_filter}
u^\ego_{t}=
\policy^\sfilter(x_t; \tilde{u}^\ego_t)\!:=
\begin{cases}
\tilde{u}^\ego_t,  & \text{if } (x_t, \tilde{u}^\ego_t) \not\in \shieldset^\ego \\
\policy^s(x_t),  & \text{if } (x_t, \tilde{u}^\ego_t) \in \shieldset^\ego.
\end{cases}
\end{equation}
Safety filter \eqref{eq:safety_filter} allows the ego agent to apply \textit{any} nominal controller $\pi_t: \reals^{n_x}\! \rightarrow \cset^\ego$ as long as
{$\big(x_t,\pi_t(x_t)\big)$ is not in the shielding set $\shieldset^\ego$;
otherwise, it overrides $\pi_t(x_t)$ with the shielding policy $\policy^s(x_t)$}.
This leads to the following result.
\begin{proposition}[Safety Filter (Prop. 1~\cite{hu2022sharp})]
\label{prop:shield}
If a set $\Omega$ is robust controlled-invariant under shielding policy $\policy^s(\cdot)$, then it is robust controlled-invariant under safety filter policy
$\policy^\sfilter\big(\,\cdot\,;\pi_t(\cdot)\big)$, for any nominal policy $\pi_t(\cdot)$.
\end{proposition}

\begin{remark}
We can replace~\eqref{eq:HRI:safety} with \emph{chance constraints} to account for uncertain inputs with unbounded supports, i.e. $\prob\left[\state \notin \failure\right] \geq 1- \delta$, where $\delta \in (0,1]$ is the tolerance level.
Then, probabilistic safety guarantees can be obtained via probabilistic shielding methods such as~\cite{bastani2021statMPS}.
\end{remark}

\subsection{Interaction-Aware SHARP}
\label{sec:SHARP:interaction}
In the SHARP framework, safety filter \eqref{eq:safety_filter} is supplied with the predicted future states and other agents' actions along each branch of the scenario tree to \emph{causally anticipate} possible future shielding events, therefore making the resulting policy aware of the (usually costly) shielding maneuvers.
However, \eqref{eq:safety_filter} involves conditioning on set inclusion relationships, making it hard to optimize within the SMPC problem.            
To this end, we modify and improve the approximate local safety filter scheme proposed in~\cite{hu2022sharp} to reformulate~\eqref{eq:safety_filter} as a convex constraint, which additionally accounts for the other agent's responses and external disturbance.

\begin{figure}[!hbtp]
  \centering
  \includegraphics[width=1.0\columnwidth]{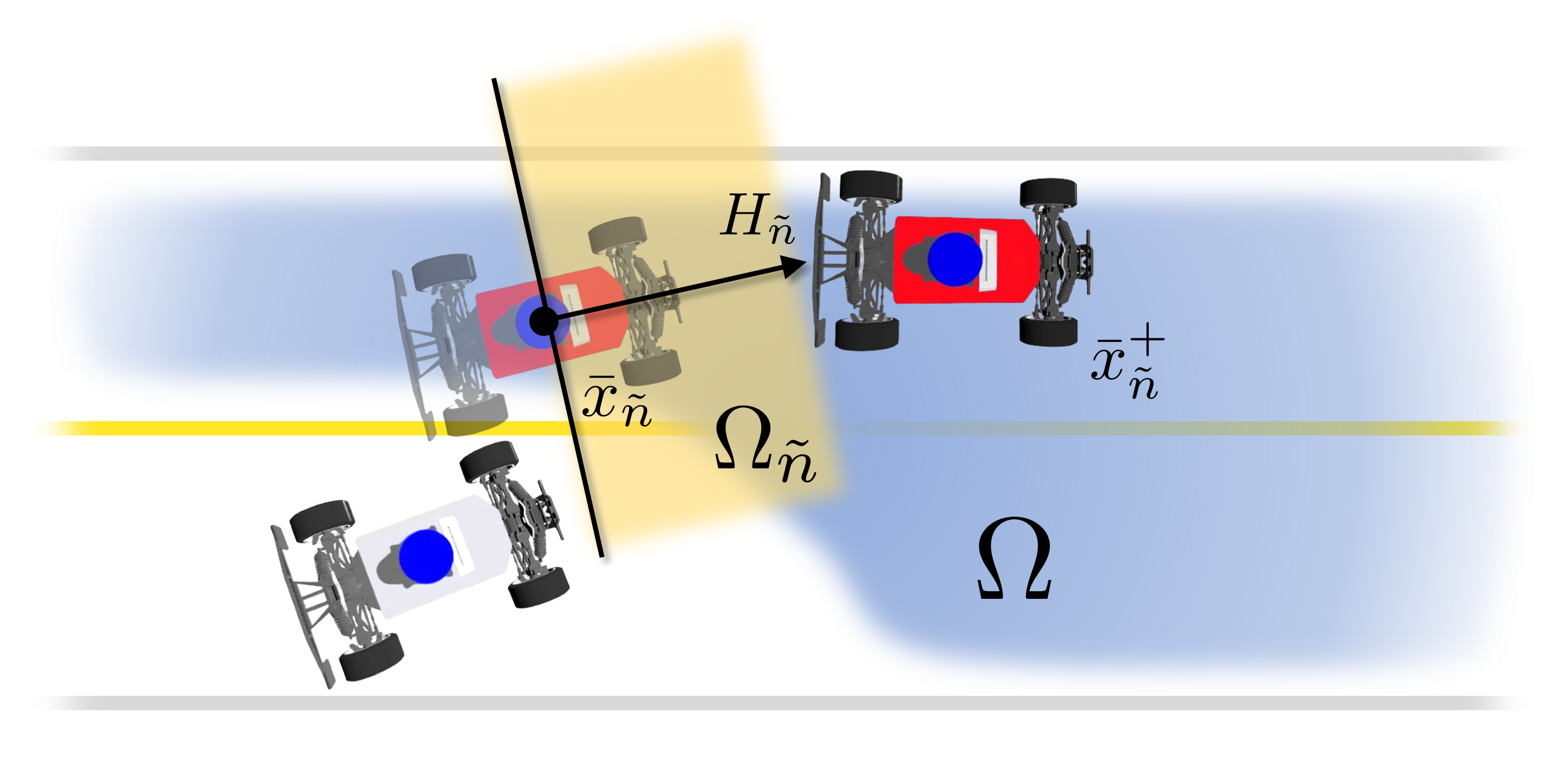}
  \caption{\label{fig:CBF} Illustration of an RCBF-based halfspace safe set $\Omega_\tnode$ (yellow), which approximates safe set $\Omega$ (blue) locally at $\state_\tnode$.
  }
\end{figure}

We start by linearizing joint system~\eqref{eq:joint_sys} at a given nominal state $\bstate_\tnode$ associated with node $\tnode$:
\begin{equation}
\label{eq:linearized_sys}
    \dstate^+_\tnode = A_\tnode \dstate_\tnode + B^\ego_\tnode \ctrl^\ego_\tnode + B^\oppo_\tnode \ctrl^\oppo_\tnode + \dstb_\tnode,
\end{equation}
where $A_\tnode := \left.{\nabla_{\state} \dyn(\state)}\right|_{\bstate_\tnode}$, $B^\ego_\tnode := B^\ego(\bstate_\tnode)$, $B^\oppo_\tnode := B^\oppo(\bstate_\tnode)$, and $\dstate_\tnode = \state_\tnode - \bstate_\tnode$.
Then, we approximate the safe set $\Omega$ as a halfspace locally at $\state_\tnode$:
\begin{equation}
\label{eq:safe_set_CBF}
\begin{aligned}
{\Omega}_{\tnode} = \{\dstate \mid {H_{\tnode}}^\top \dstate \geq 0 \},
\end{aligned}
\end{equation}
where $H_{\tnode} := \bstate^+_\tnode - \bstate_{\tnode}$ approximates the normal vector of the tangent space of $\Omega$ at $\state_{\tnode}$, as illustrated in Figure~\ref{fig:CBF}.
Now, given uncertain linear system~\eqref{eq:linearized_sys} and halfspace safe set $\Omega_\tnode$, we extend the result in~\cite{agrawal2017discrete} to construct an affine robust control barrier function (RCBF) constraint that approximates~\eqref{eq:safety_filter}.
This RCBF locally captures the shielding maneuver by inducing a constraint that certifies the controlled invariance of the approximate safe set ${\Omega}_{\tnode}$, which is formalized in Lemma~\ref{lem:CBF}.

\begin{definition}[Discrete-Time Exponential RCBF]
\label{def:RCBF}
A map $h(\state)$ is a discrete-time exponential robust control barrier function for system $\state_{t+1} = \dyn(\state_t, \ctrl_t, \dstb_t)$ if:
\begin{enumerate}[(a)]
    \item $h(\state_0) \geq 0$ and,
    \item $\exists \ctrl_t \in \cset$ such that $h(\state_{t+1}) - h(\state_t) + \gamma h(\state_t) \geq 0$ for all $t \geq 0$, $\dstb_t \in \dset$, and some $\gamma \in (0, 1]$.
\end{enumerate}
\end{definition}

\begin{lemma}
\label{lem:CBF}
Let Assumption~\ref{assump:dset} hold. For uncertain linear system~\eqref{eq:linearized_sys}, given a control $\ctrl^\oppo \in \cset^\oppo$ of the other agent, set ${\Omega}_{\tnode}$ define by \eqref{eq:safe_set_CBF} is robust controlled-invariant under ego control $\ctrl_\tnode^\ego \in \cset^\ego$ if it satisfies:
\begin{equation}
\label{eq:cvx_sh_constr}
\begin{aligned}
   &H_{\tnode}^\top \left[ \left(A_{\tnode} + (\gamma-1)I \right) \dstate_\tnode + B^\ego_{\tnode} \ctrl_\tnode^\ego + B^\oppo_{\tnode} \ctrl^\oppo + d^*_{\tnode} \right] \\
   & =: K_{\tnode}(\dstate_\tnode, \ctrl_\tnode^\ego, \ctrl^\oppo) \geq 0,
\end{aligned}
\end{equation}
for all $\delta_x \in {\Omega}_{\tnode}$ and some $\gamma \in (0, 1]$.
Here, the optimal (worst-case) disturbance $\dstb^*_\tnode$ is defined element-wise as:
\begin{equation}
\label{eq:cvx_sh_constr:opt_dstb}
\dstb^{*,i}_\tnode\!:=
\begin{cases}
\bar{\dstb}^i_{\min},  & \text{if } H_\tnode^i \geq 0 \\
\bar{\dstb}^i_{\max},  & \text{otherwise}.
\end{cases}
\end{equation}
\end{lemma}

\noindent The proof can be found in Appendix~\ref{apdx:proof:CBF}. Inequality~\eqref{eq:cvx_sh_constr} is affine (and hence \emph{convex}) in $\dstate_\tnode$ and $\ctrl^\ego_{\tnode}$, and $\ctrl^\oppo$, which is efficient to optimize and can be infused into the ST-SMPC problem as an inequality constraint for causally \emph{predicting} future shielding events.
Instead of using the worst-case action in~\eqref{eq:cvx_sh_constr}, we allow it to depend on the other agent's \emph{predicted} action $\ctrl^\oppo_{\tnode}$, which is a decision variable affected by the ego's action and state of the system.
This enables SHARP to account for interaction while predicting shielding events.
At time $t$, based on the last optimized scenario tree $\nodeset^*_{t-1}$, we can perform a one-step simulation rollout per~\eqref{eq:safety_filter} to identify nodes at which the shielding policy shall be used.
We denote the set of all shielding nodes at time $t$ as $\nodeset^\shield_t$.
The convex shielding-aware constraint~\eqref{eq:cvx_sh_constr} is enforced for all $\tnode^\shield \in \nodeset^\shield$.
The procedure of finding  $\nodeset^\shield_t$ is summarized in Algorithm~\ref{alg:SHARP}.

\begin{algorithm}[!hbtp]
	\caption{Identify shielding nodes}
	\label{alg:SHARP}
	\begin{algorithmic}[1]
	\Require Last optimized scenario tree defined by node set $\nodeset^*_{t-1}$, scenario tree defined by node set $\nodeset_t$ 
	\Ensure Shielding node set $\nodeset^\shield_t$
    \State Initialize the shielding node set: $\nodeset^\shield_t \gets \emptyset$
    \ForAll{$(\tnode^*, \tnode) \gets \texttt{zip}(\nodeset^*_{t-1}, \nodeset_{t})$}
    	\State $\tilde{\state}_{\tnode^*}^+ \gets$ one-step forward simulation using $\state_{\tnode^*}$, $\ctrl^\ego_{\tnode^*}$, $\ctrl^\oppo_{\tnode^*}$, $\bar{\dstb}^M_{\tnode^*}$ and dynamics~\eqref{eq:ST-SMPC:dyn}
    	\If{$\tilde{\state}_{\tnode^*}^+ \notin \Omega$}
    	    \State Compute $H_{\tnode}$, $A_{\tnode}$, $B^\ego_{\tnode}$, $B^\oppo_{\tnode}$, and $d^*_{\tnode}$ according to Section~\ref{sec:SHARP:interaction}
    	    \State $\nodeset^\shield_t \gets \nodeset^\shield_t \cup \{\tnode\}$
    	\EndIf
	\EndFor
	\end{algorithmic}
\end{algorithm}

\subsection{Overall Algorithmic Approach }
\label{sec:SMPC}
Given the current state measurement $\hat{\state}_t$, the last updated belief state $\hat{\bel}_t$, and a scenario tree defined by node sets $\nodeset_t$, we can reformulate~\eqref{eq:HRI} as an ST-SMPC problem in a similar format to~\cite{bernardini2011stabilizing}:
\begin{subequations}
\label{eq:ST-SMPC}
\begin{align}
\min_{\substack{\mathbf{U}^\ego_t}} \ \ 
&\label{eq:ST-SMPC:cost} \sum_{\tnode \in \nodeset_t \setminus \nodesetleaf_t} \prob_{\tnode} \ell (\state_\tnode, \ctrl^\ego_\tnode) + \sum_{\tnode \in \nodesetleaf_t} \prob_{\tnode} \ell_F (\state_\tnode) \\
\label{eq:ST-SMPC:IC} \text{s.t.} \ \  &\state_{\node_0}=\hat{\state}_{t}, \ \bel_{\node_0}=\hat{\bel}_{t}, \\
&\label{eq:ST-SMPC:dyn} \state_{\tnode} =  \dyn (\state_{\pre{\tnode}}) + B^\oppo(\state_{\pre{\tnode}}) \ctrl^\oppo_{\tnode}(\state_{\pre{\tnode}}, \ctrl^\ego_{\pre{\tnode}}) + \notag \\
&\hspace{0.5cm} B^\ego(\state_{\pre{\tnode}}) \ctrl^\ego_{\pre{\tnode}} + \bar{\dstb}^{M}_{\tnode}, &&\hspace{-2.9cm}\forall \tnode \in \nodeset_t \setminus \{n_0\}  \\
&\label{eq:ST-SMPC:b_dyn} {\bel}_{\tnode}=\tilde{g}\left({\bel}_{\pre{\tnode}}, \state_{\tnode}, \ctrl^\ego_{\pre{\tnode}}\right),  &&\hspace{-2.9cm}\forall \tnode \in \nodeset_t^d\\
&\label{eq:ST-SMPC:b_dyn_trans} {\bel}_{\tnode}=g^t\left({\bel}_{\pre{\tnode}}\right), &&\hspace{-2.9cm} \forall \tnode \in \nodeset_t^e \\
&\label{eq:ST-SMPC:u_constr} \ctrl^\ego_\tnode \in \cset^\ego,~\ctrl^\oppo_{{\tnode}} \in \cset^\oppo, &&\hspace{-2.9cm}\forall \tnode \in \nodeset_t \setminus \nodesetleaf_t\\
&\label{eq:ST-SMPC:SHARP} K_{\tnode}(\dstate_\tnode, \ctrl_\tnode^\ego, \ctrl_\tnode^\oppo) \geq 0, &&\hspace{-2.9cm}\forall \tnode \in \nodeset^\shield_t
\end{align}
\end{subequations}
where $\nodesetleaf_t$ is the set of all leaf nodes,
(i.e. ones that do not have a descendant),
$\nodeset^d_t$ and $\nodeset^e_t$ are the set of dual control and exploitation nodes, respectively, set $\mathbf{U}^\ego_t := \{ \ctrl^\ego_\tnode \in \reals^{m_\ego}: \tnode \in \nodeset_t \setminus \nodesetleaf_t \}$ is the collection of the ego's control inputs associated with all non-leaf nodes, and $\nodeset^\shield_t$ is shielding node set returned by Algorithm~\ref{alg:SHARP}, containing nodes at which the convex shielding-aware constraint~\eqref{eq:cvx_sh_constr} is imposed.
Objective function~\eqref{eq:ST-SMPC:cost} approximates~\eqref{eq:HRI:obj} based on uncertainty samples of the scenario tree.
Constraints~\eqref{eq:ST-SMPC:IC}-\eqref{eq:ST-SMPC:SHARP} capture the initial (belief) state, dynamics of the physical states, belief dynamics of the exploration and exploitation steps, control limits, and convex shielding-aware constraints in~\eqref{eq:cvx_sh_constr}, respectively.
Problem~\eqref{eq:ST-SMPC} is a nonconvex trajectory optimization problem, which can be solved using general-purpose nonconvex solvers such as SNOPT~\cite{gill2005snopt} and IPOPT~\cite{wachter2006implementation}.
The optimal solution $\mathbf{U}^{\ego,*}_t$ to \eqref{eq:ST-SMPC} is implemented in a receding horizon fashion, i.e. $\policy^\ego_{\text{IDSMPC-SHARP}}(\hat{\state}_t, \hat{\bel}_t) := \ctrl_{\node_0}^{\ego,*}$, and we refer to this as the shielding-aware implicit dual SMPC (IDSMPC-SHARP) policy. 
Our overall algorithmic approach, which is centered around the IDSMPC-SHARP policy, can be found in Algorithm~\ref{alg:overall}.

\begin{remark}
\label{rmk:cvx_sh_constr}
The sole purpose of incorporating RCBF constraint~\eqref{eq:cvx_sh_constr}, which approximates safety filter policy~\eqref{eq:safety_filter}, in ST-SMPC~\eqref{eq:ST-SMPC:SHARP} is to \emph{predict} future shielding overrides, thus proactively improving the long-term planning performance.
This approximation does \emph{not} affect the recursive safety of the system, which is guaranteed through the shielding step (Line 6 in Alg.~\ref{alg:overall}, proven in Theorem~\ref{thm:safety}).
\end{remark}

\begin{remark}
When cost functions in~\eqref{eq:ST-SMPC:cost} are quadratic, i.e. $\ell(\state, \ctrl) = \|\state\|^2_Q + \|\ctrl\|^2_R$ and $\ell^\ego_F(\state)=\|\state\|^2_{Q_F}$, an approximation technique can be deployed to analytically evaluate the expected cost with respect to the state.
Let $A_{\pre{\tnode}}$ denote the Jacobian of dynamics~\eqref{eq:ST-SMPC:dyn} evaluated at the mean value of state $\mu^\state_{\pre{\tnode}}$ and ego's control $\ctrl^\ego_{\pre{\tnode}}$.
Then, the state will remain Gaussian-distributed along each branch in the scenario tree, whose covariance is given by the recursive formula: $\covar^\state_{\tnode} = \covar^{\bar{\dstb}}_{\tnode} (\mu^\state_{\pre{\tnode}}, \ctrl^\ego_{\pre{\tnode}}) + A_{\pre{\tnode}} \covar^\state_{\pre{\tnode}} A^\top_{\pre{\tnode}}$.
We recall that $\expectation_{\xi \sim \gaussian(0, I)} \left[ (W\xi)^\top Q (W\xi) \right] = \trace(W^\top Q W)$.
Therefore, objective~\eqref{eq:ST-SMPC:cost} can be approximated as
\begin{equation*}
\begin{aligned}
\sum_{\tnode \in \nodeset_t \setminus \nodesetleaf_t} \prob_{\tnode} \ell (\state_\tnode, \ctrl^\ego_\tnode) + \sum_{\tnode \in \nodesetleaf_t} \prob_{\tnode} \ell_F (\state_\tnode) + \sum_{\tnode \in \nodeset_t} \trace(Q\covar^\state_{\tnode}) 
\end{aligned}
\end{equation*}
where the path transition probability of node $\tnode$ evaluates to $\prob_{\tnode} = \prob(\theta^M_n \mid \ivec_{\pre{\node}}; M_\node) \prob\left(M_\node \mid \ivec_{\pre{\node}}\right) \prob_{\pre{\tnode}}$.
This way, only the hidden states are sampled for computing the expected cost, and no $\bar{\dstb}^M$ sample is needed.
\end{remark}

\begin{algorithm}[!hbtp]
	\caption{Shielding-aware safe dual control}
	\label{alg:overall}
	\begin{algorithmic}[1]
	\Require Initial state $\state_0$, initial belief state $\bel_0$, Q-value functions $\qfunc_i^M(\cdot)$ in~\eqref{eq:exo-agent_ctrl_model}, shielding mechanism $(\Omega, \policy^s)$
    \State Initialization: $t \gets 0$, $\nodeset^*_0 \gets \emptyset$
    \While{Planning goal is not reached}
        \LineComment{Planning:}
    	\State $\nodeset_t \gets$ construct a scenario tree using Algorithm~\ref{alg:tree}
    	\State $\nodeset^\shield_t \gets$ identify the shielding nodes and construct shielding-aware constraints using Algorithm~\ref{alg:SHARP}
    	\State $\policy^\ego_{\text{IDSMPC-SHARP}}(\state_t, \bel_t),~\nodeset^*_t \gets$ solve ST-SMPC~\eqref{eq:ST-SMPC}
    	\LineComment{Shielding:}
    	\State Apply safety filter policy~\eqref{eq:safety_filter} to the ego agent: $u^\ego_{t} = \policy^\sfilter(x_t; \policy^\ego_{\text{IDSMPC-SHARP}}(\state_t, \bel_t))$
    	\LineComment{Belief Update:}
    	\State Measure new state $\state_{t+1}$ 
    	\State $\bel_{t+1} \gets$ update belief states using~\eqref{eq:Bayes_est_meas_theta}-\eqref{eq:Bayes_est_time} 
    	\State Update time: $t \gets t+1$
    \EndWhile
	\end{algorithmic}
\end{algorithm}

\begin{figure*}[!hbtp]
  \centering
  \includegraphics[width=1.8\columnwidth]{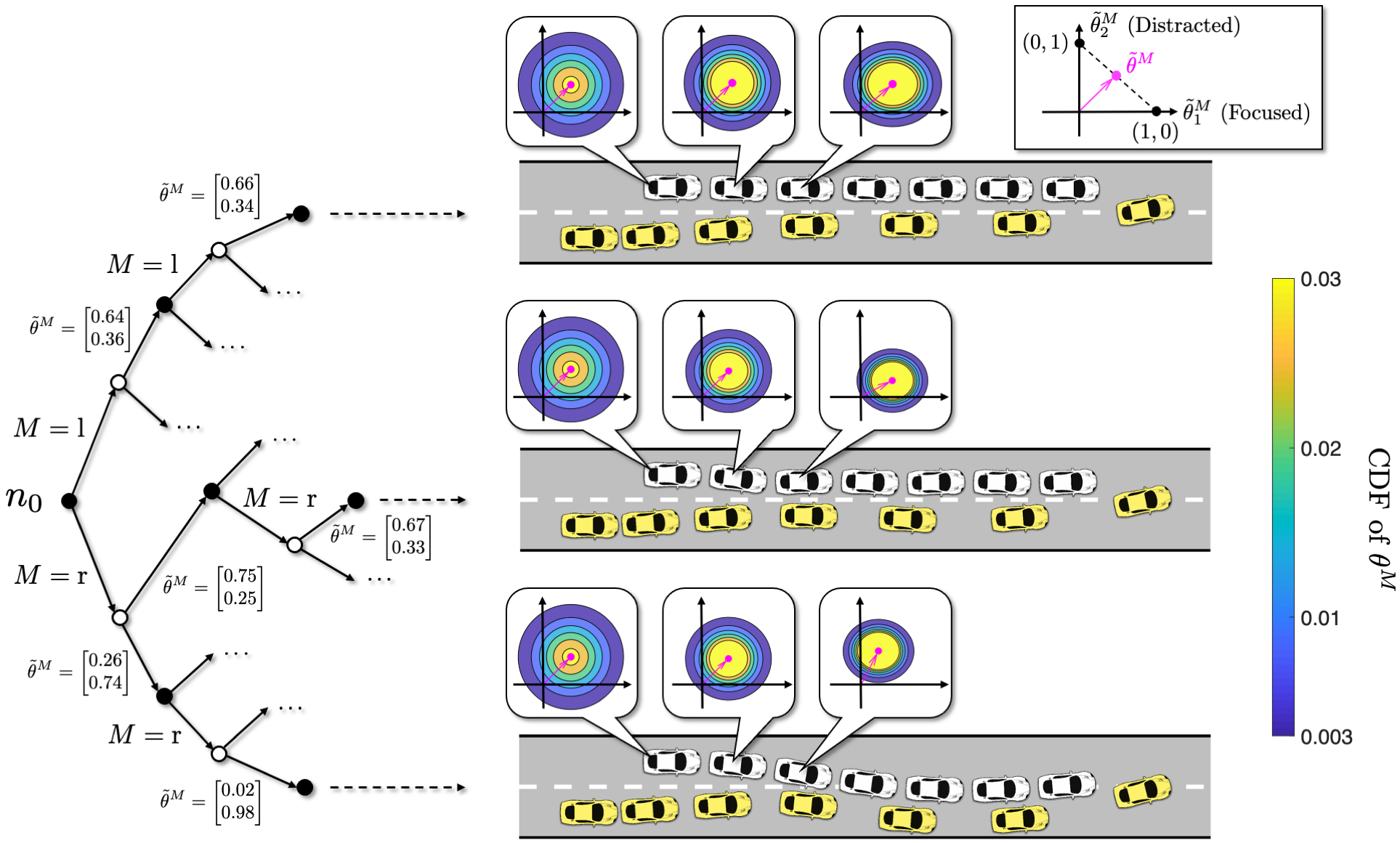}
  \caption{\label{fig:stree_sim} Illustration of a scenario tree and the resulting optimized scenario trajectories with $N^d = 2$ dual control time steps and $N^e = 4$ exploitation steps for the highway overtaking example in Section~\ref{sec:sim:Ex1}.
  The human-driven vehicle and autonomous car are plotted in white and yellow, respectively.
  Hidden state $\theta^M$ is modeled as a 2D Gaussian random variable, which is normalized to a 1-simplex for visualization.
  A white circle $\circ$ denotes an intermediate node partially determined by an $M$ sample, and a black circle $\bullet$ denotes a node fully determined by both $M_\tnode$ and $\tilde{\theta}^M$ samples.
  The magenta arrows show the MAP mean of the normalized hidden state $\tilde{\theta}^{M}$.
  The contour plots display level sets of the cumulative distribution function (CDF) of~$\theta^{M}$.
  Uncertainty is \emph{less} significantly reduced in the case where the human prefers the left lane (upper branch) since their behavior is less influenced by the robot's (probing) actions than in the right lane case.
  }
\end{figure*}

\subsection{Properties of the Planning Framework}

In this section, we examine the properties of the proposed IDSMPC-SHARP control policy, as well as the behavior of the joint system~\eqref{eq:joint_sys} in closed-loop with Algorithm~\ref{alg:overall}.

\begin{theorem}[Dual Control Effect]
\label{thm:dual_control}
The feedback control policy $\policy^\ego_{\text{IDSMPC-SHARP}}(\cdot,\cdot)$ obtained by solving~\eqref{eq:ST-SMPC} produces dual control effect.
\end{theorem}

\begin{proof}
From Lemma~\ref{lem:theta_update}, for any given time $t \geq 0$, the robot's control $\ctrl^\ego_t$ can affect $\covar^{\theta^M_{-}}_{t+1}$, the covariance (second-order moment) of the belief over~$\theta^M$.
Therefore, the policy $\policy^\ego_{\text{IDSMPC-SHARP}}$ produces dual control effect for hidden state $\theta^M$ per Definition~\ref{def:DC}\ref{def:DC_a}.
We recall the measurement update equation for the categorical belief over $M$ from~\eqref{eq:Bayes_est_meas_M}:
$$\distr(M_{-} \mid \ivec_{t+1}) = \frac{\distr(\state_{t+1} \mid \ctrl^\ego_t, \ivec_t; M) \distr(M \mid \ivec_{t}) }{ \distr(\state_{t+1} \mid \ctrl^\ego_t, \ivec_t)},$$
which shows that the ego's control $\ctrl^\ego_t$ can affect all components of the categorical distribution over $M$, and thereby its entropy:
$$H(M_{-} \mid \ivec_{t+1}) \propto \sum_{\tilde{M} \in \mset} \prob(\tilde{M} \mid \ivec_{t+1}) \log \prob(\tilde{M} \mid \ivec_{t+1}),$$
implying dual control effect for $M$ per Definition~\ref{def:DC}\ref{def:DC_b}.
\qed
\end{proof}

\begin{remark}[Optimality]
The solution to ST-SMPC~\eqref{eq:ST-SMPC} is in general sub-optimal with respect to Bellman recursion~\eqref{eq:sto_DP} due to the approximate belief state dynamics $\tilde{g}$, expected cost approximated with uncertainty samples, truncated exploration steps, and that the solution to the nonconvex program~\eqref{eq:ST-SMPC} is oftentimes only locally optimal.
Therefore, the optimal exploration-exploitation trade-off is generally not achieved.
Nonetheless, since the ego's control produces dual control effect for the hidden states per Theorem~\ref{thm:dual_control}, it thereby automatically balances the nominal planning performance and uncertainty reduction, to the extent that local optimality of~\eqref{eq:ST-SMPC} is achieved.
\end{remark}

\begin{remark}[Guaranteeing Feasibility]
ST-SMPC~\eqref{eq:ST-SMPC} is feasible as long as the convex shielding-aware constraint~\eqref{eq:ST-SMPC:SHARP} is satisfied.
As pointed out in~\cite{agrawal2017discrete}, it is not guaranteed that there exists a feasible solution to constraint~\eqref{eq:ST-SMPC:SHARP} when the ego's control input is bounded.
In order to ensure that~\eqref{eq:ST-SMPC} is recursively feasible, we may incorporate~\eqref{eq:ST-SMPC:SHARP} as a soft constraint with a slack variable, i.e. $K_{\tnode}(\dstate_\tnode, \ctrl_\tnode^\ego, \ctrl_\tnode^\oppo) \geq s_\tnode,~s_\tnode \leq 0$.
Note that relaxing constraint~\eqref{eq:ST-SMPC:SHARP} does \emph{not} affect recursive safety, which is enforced by the shielding step (Line 6 in Alg.~\ref{alg:overall}) and proven in Theorem~\ref{thm:safety}.
\end{remark}

\begin{theorem}[Recursive Safety]\label{thm:safety}
Suppose the initial state is in the safe set, i.e. $\state_0 \in \Omega$, then the joint system~\eqref{eq:joint_sys} in closed-loop with Algorithm~\ref{alg:overall} remains safe, i.e. $\state_t \in \Omega,~\forall t >0.$
\end{theorem}

\begin{proof}
The result is a direct consequence of Proposition~\ref{prop:shield} and Definition~\ref{def:RCI_set}.
\qed
\end{proof}

\section{Simulation Studies}
\label{sec:sim}

\subsection{Simulation Setup}
\label{sec:sim:setup}
We evaluate our proposed implicit dual scenario tree--based SMPC (IDSMPC, given by solving~\eqref{eq:ST-SMPC} without the shielding-aware constraint~\eqref{eq:ST-SMPC:SHARP}) and IDSMPC-SHARP (given by solving~\eqref{eq:ST-SMPC}) planners on simulated driving scenarios.
In both planning and simulation, vehicle and pedestrian dynamics are described by the 4D kinematic bicycle model in~\cite{zhang2020optimization} and the 4D unicycle model in~\cite{fridovich2020efficient}, respectively, both discretized with a time step of $\Delta t =$ 0.2 s.
All simulations are performed using MATLAB and YALMIP~\cite{lofberg2004yalmip} on a desktop with an Intel Core i7-10700K CPU.
All nonlinear MPC problems are solved with SNOPT~\cite{gill2005snopt}.
Parameter values used for simulation can be found in Table~\ref{tab:planner_param_sim}.
The open-source code is available online.\footnote{\scriptsize \url{https://github.com/SafeRoboticsLab/Dual_Control_HRI}}

\subsubsection{Baselines.}
We compare our proposed IDSMPC and IDSMPC-SHARP planners against four baselines:
\begin{itemize}
    \item Explicit dual SMPC (EDSMPC), which augments the stage cost $\ell$ in \eqref{eq:HRI:obj} with an information gain term $\lambda (\entropy(\bel_{k})-\entropy(\bel_{k+1}))$ proposed by~\cite{sadigh2018planning}, leading to an explicit dual stochastic optimal control problem.
    Here, $\lambda > 0$ is a fine-tuned weighting factor.
    Similar to IDSMPC, the expected cost is also computed approximately using uncertainty samples.
    
    \item Non-dual scenario-based SMPC (NDSMPC), which is based on solving~\eqref{eq:HRI} with a scenario tree that does not propagate belief states with the measurement update~\eqref{eq:ST-SMPC:b_dyn} (so the resulting policy does not have dual control effect).
    A similar scenario program can also be found in in~\cite{bernardini2011stabilizing,schildbach2015scenario,hu2022sharp}.
    
    \item Certainty-equivalent MPC (CEMPC), which is based on solving~\eqref{eq:HRI} with the certainty-equivalence principle~\cite{mesbah2018stochastic,arcari2020dual}.
    The maximum a posteriori (MAP) strategy alignment method~\cite{peters2020inference} is used by the ego agent to modify the planning problem, e.g. in Example~\ref{example:1}, if both the ego and other vehicles drive on the inner lane, and the MAP estimate of $M$ is $\text{NY}$ (not yielding), then the reference lane of the ego would be set to the outer lane (by modifying costs $\ell(\cdot)$ and $\ell_F(\cdot)$) so that it can safely overtake the other agent.
    
    \item Iterative linear-quadratic game with inference-based strategy alignment (ISA-iLQ), proposed by~\cite{peters2020inference} and originally developed in~\cite{fridovich2020efficient}.
    An input projector is used to enforce $\ctrl^\oppo \in \cset^\oppo$ and $\ctrl^\ego \in \cset^\ego$.
\end{itemize}
EDSMPC is a dual control planner while the other three do not generate dual control effect.
All planners use the same quadratic cost functions $\ell$ and $\ell_F$ for penalizing reference tracking error and control magnitude, and are equipped with the same Bayesian-inference-based method for inferring the other agent's intent.
To incrementally interpret the results, IDSMPC and four baselines are unshielded and, to account for safety, we use the soft constrained MPC approach in~\cite{zeilinger2014soft}, which relaxes the original hard constraints $\state_\tnode \notin \failure$ with slack variables for each node $\tnode \in \nodeset_t$, making those planners \emph{safety-aware but shielding-agnostic}.
IDSMPC-SHARP uses a shielding mechanism given by a contingency planner design~\cite{hardy2013contingency, bajcsy2021analyzing} that safeguards against the other agent's behavioral uncertainties and external disturbances (under Assumption~\ref{assump:dset}).
The contingency planning problem was solved online with the iterative Linear-Quadratic Regulator (iLQR) method~\cite{todorov2005generalized}.
We use the model predictive shielding (MPS) algorithm~\cite{bastani2021safe} to ensure the recursive feasibility of the shielding policy.

\begin{figure*}[!hbtp]
     \centering
     \includegraphics[width=2.0\columnwidth]{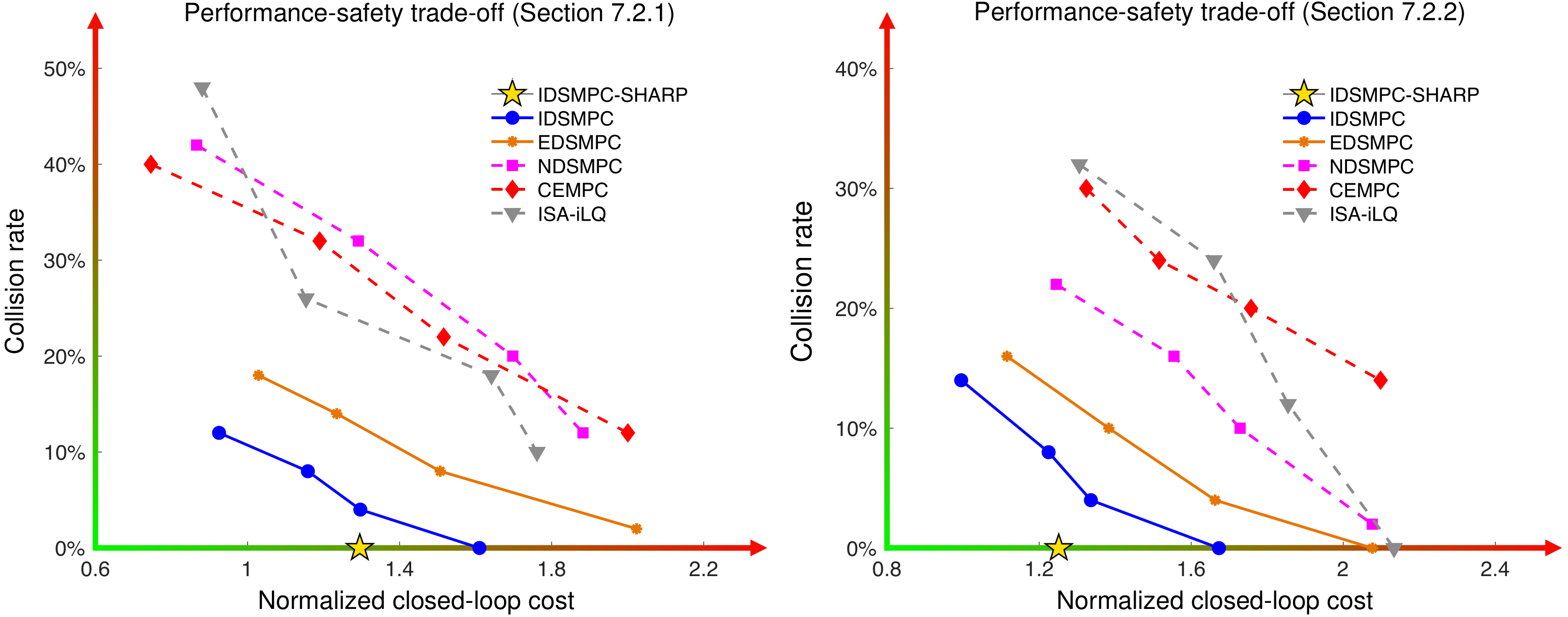}
     \caption{\label{fig:Ex1_tradeoff} Performance-safety trade-off of simulation examples in Section~\ref{sec:sim:Ex1} and~\ref{sec:sim:Ex2}.
     Closed-loop cost $J^\ego_\cl$ is normalized by $1 \times 10^4$.
     For the unshielded planners (IDSMPC, EDSMPC, NDSMPC, CEMPC, and ISA-iLQ), each data point is obtained based on a distinct set of safety-critical tuning parameters of 50 trials with different random seeds.
     The data point associated with 50 simulation trials using IDSMPC-SHARP is shown in a yellow star.
     Curves of the dual control policies are shown as solid lines while those of the non-dual ones are shown as dashed lines.
     IDSMPC outperforms all baselines in terms of overall performance-safety trade-off.
     Due to shielding and the SHARP framework, IDSMPC-SHARP yields zero collision rate and about $20\%$ less cost compared to the safest IDSMPC design.
     }
\end{figure*}

\subsubsection{Metrics.}
To measure the planning performance, we consider the following two metrics:
\begin{itemize}
    \item Closed-loop cost, defined as $J^\ego_\cl := \sum_{t=0}^{T_\Sim} \ell(\state_t, \ctrl_t^\ego)$, where $T_\Sim$ is the simulation horizon, and $\state_{[0:T_\Sim]}$, $\ctrl_{[0:T_\Sim]}$ are the \emph{executed} trajectories (with replanning).
    \item Collision rate, defined as $N_{\text{coll}} / N_{\text{trial}} \times 100\%$, where $N_{\text{coll}}$ is the number of trials that a collision happens, i.e. $x_t \in \failure$,  and $N_{\text{trial}}$ is the total number of trials.
\end{itemize}

\subsubsection{Hypotheses.}
\label{subsec:hypotheses}
We make three hypotheses, which are confirmed by our simulation results.
\begin{itemize}
    \item \textbf{H1 (Performance and Safety Trade-off).} \emph{Dual control planners result in a better performance-safety trade-off than non-dual baselines.}
    
    \item \textbf{H2 (Implicit vs Explicit Dual Control).} \emph{Explicit dual control is less efficient than its implicit counterpart, even with fine tuning.}
    
    \item \textbf{H3 (Improved efficiency with SHARP).} \emph{Planning efficiency is improved using IDSMPC-SHARP compared to that of IDSMPC, even fine-tuned for safety.}
\end{itemize}

\begin{figure}[!hbtp]
  \centering
  \includegraphics[width=1.0\columnwidth]{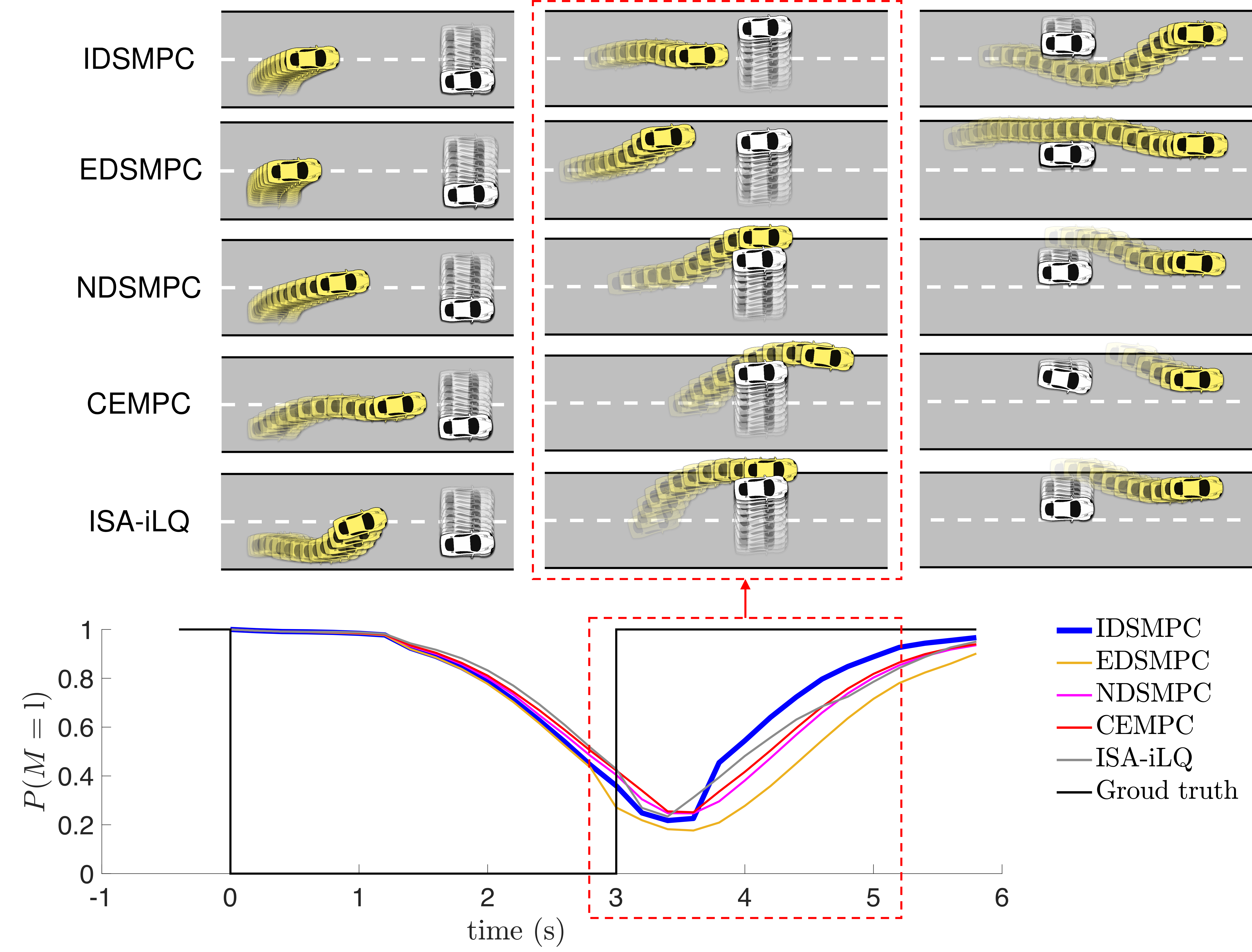}
  \caption{\label{fig:Ex1_Trajs} Simulation snapshots of the highway driving example presented in Section~\ref{sec:sim:Ex1}, where the ego vehicle in yellow seeks to overtake the human-driven vehicle in white.
  Longitudinal positions are shown in relative coordinates with $p_x^H = 0$. The left, middle, and right columns display trajectories for $t=[0,3]$ s, $t=[3,5]$ s, and the remainder of the trajectories.
  The bottom figure shows $\prob(M=\text{l})$ for all four planners over time.
  Our proposed IDSMPC planner yielded a clean and safe overtaking maneuver of the robot while the non-dual planners led to unsafe trajectories.
  When the robot used EDSMPC, it was stuck in a narrow window left to the human-driven car, resulting in a less efficient trajectory.
  }
\end{figure}
\begin{figure*}[!hbtp]
  \centering
  \includegraphics[width=1.8\columnwidth]{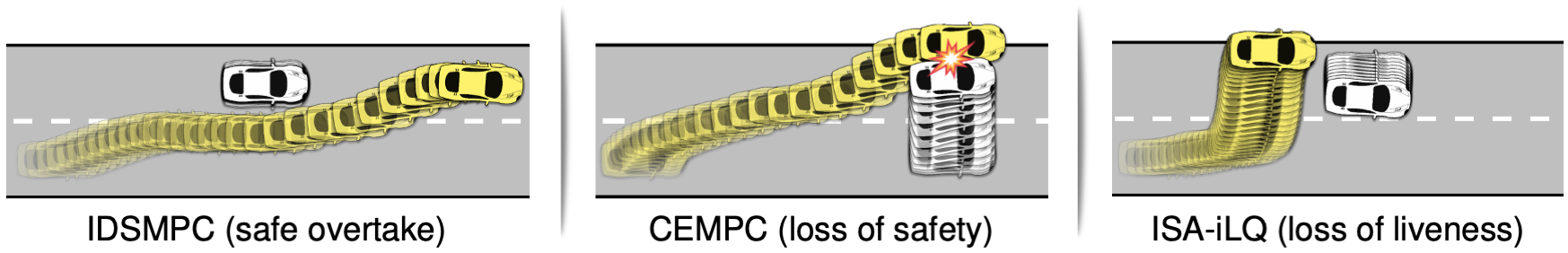}
  \caption{\label{fig:Ex1_safety_liveness} One trial of the highway driving example in Section~\ref{sec:sim:Ex1}.
  The left figure shows that the robot (yellow) successfully overtook the human-driven vehicle (white) in 6 s using IDSMPC.
  The middle and right figures display an unsafe trajectory for $t=[0,4.6]$ s using CEMPC, and a trajectory generated where the robot failed to overtake the human in 10 s using ISA-iLQ.
  }
\end{figure*}
\subsection{Simulated Agents}

\subsubsection{Agent's Policy.}
\label{sec:sim:oppo_policy}
To show the efficacy of our method in general interaction planning settings, we produce the other agent's motion using an optimization-based simulator similar to~\cite{baseggio2011mpc, guo2013understanding}.
The design parameters of the simulator are not accessible to the ego agent.
While we have made our best effort to produce plausible simulated human agent behaviors that do not fall into the hypotheses captured by the motion prediction model used by ST-SMPC~\eqref{eq:ST-SMPC}, we acknowledge that our simulated human behavior might still differ from real-world human data, which are usually expensive and difficult to obtain, especially in an interactive setting.
Nonetheless, we show in Section~\ref{sec:sim:waymo} an example where the human's trajectories are from the Waymo Open Motion Dataset~\cite{sun2020scalability}.

\subsubsection{Objective and Awareness Uncertainty.}
\label{sec:sim:Ex1}
Consider a highway driving scenario depicted in Figure~\ref{fig:Ex1_Trajs}, involving an autonomous vehicle (ego, colored in yellow), which is tasked to overtake a human-driven vehicle (the other agent, colored in white).
The continuous hidden state is defined as $\theta^M := (\theta^M_{D}, \theta^M_{F})$ where $\theta^M_D$ and $\theta^M_F$ capture the level of distraction and focus of the human, respectively.
A focused human accounts for the safety of the joint system (e.g. avoiding and making room for the robot when it attempts to merge in front of the human), while a distracted human does not.
The discrete hidden state $M$ models if the human prefers to drive in the left lane or in the right lane, i.e. $M \in \{\text{l}, \text{r}\}$. 
An optimized scenario tree of this example obtained by solving IDSMPC is visualized in Figure~\ref{fig:stree_sim}.
\begin{figure*}[!hbtp]
  \centering
  \includegraphics[width=1.9\columnwidth]{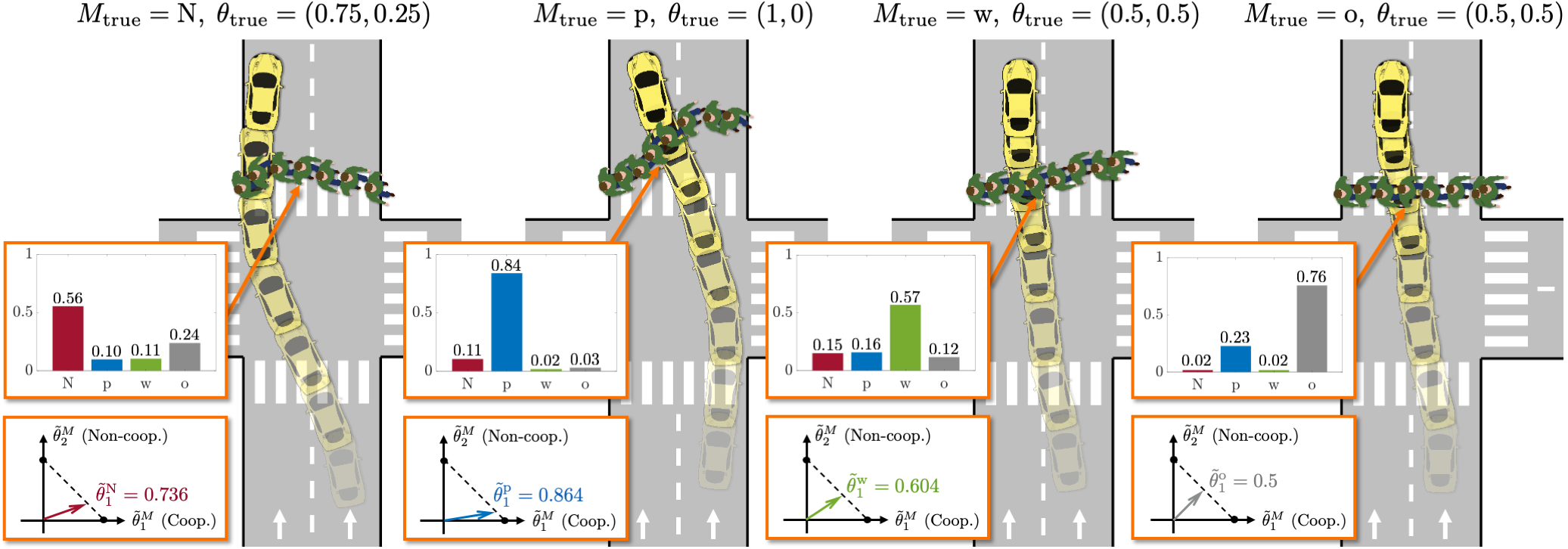}
  \caption{\label{fig:Ex2} Simulation snapshots of Example 2 using the proposed IDSMPC planner. 
  The human's mode is fixed throughout the simulation.
  In the orange blocks we display at $t =$ 2.6 s the robot's running belief $\distr(M \mid \ivec_{t})$ and $\tilde{\theta}^M$, which is the MAP mean of $\theta^M$ normalized to a 1-simplex.
  The robot was able to quickly identify the human's hidden states and planned a collision-free trajectory in all four trials, accounting for the anticipated uncertainty reduction and interactions with the human.
  }
\end{figure*}
\begin{figure*}[!hbtp]
  \centering
  \includegraphics[width=1.7\columnwidth]{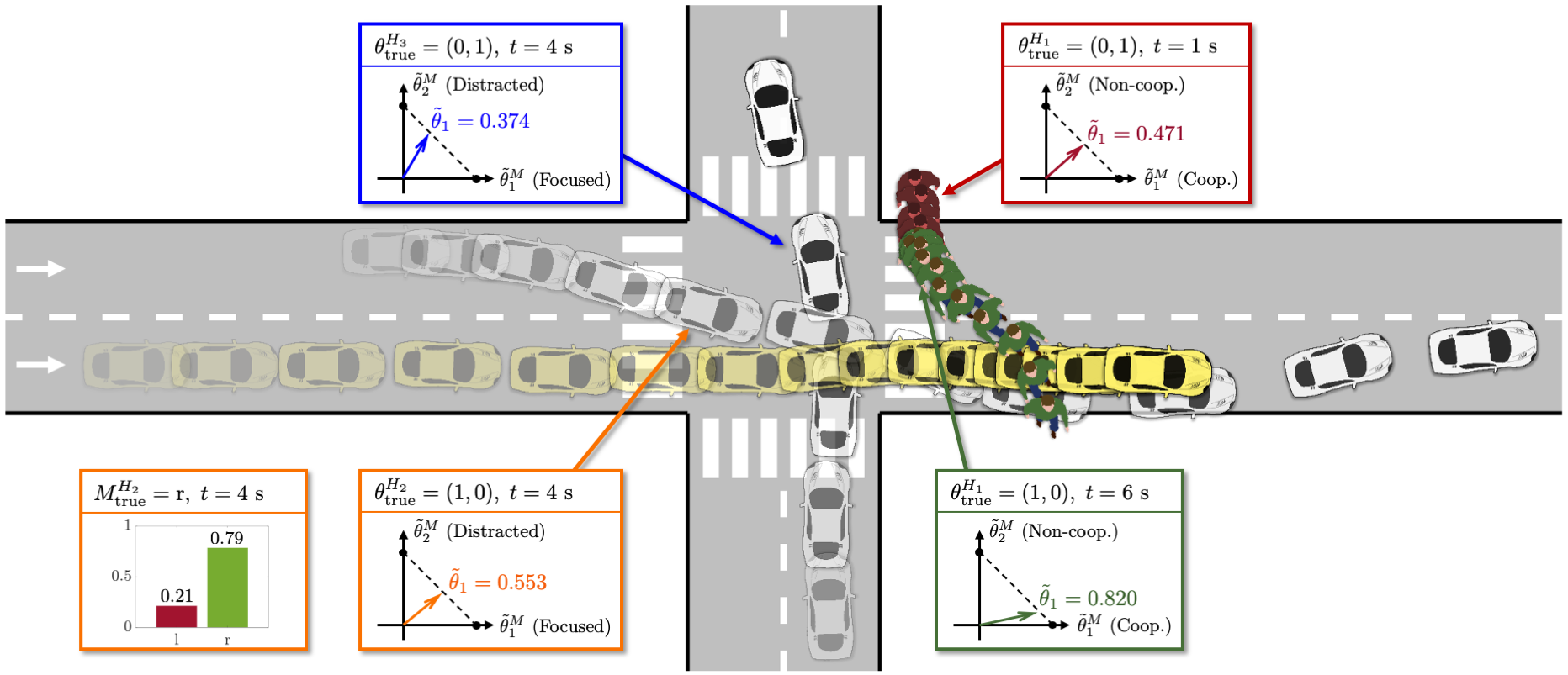}
  \caption{\label{fig:Ex3} Simulation snapshots and estimated hidden states of a multi-agent interaction scenario with a pedestrian ($H_1$) and two human-driven vehicles ($H_2$ and $H_3$) using the IDSMPC planner.
  In the boxes we display at $t =$ 4 s the continuous hidden state estimate $\tilde{\theta}^M$ (normalized to a 1-simplex) of other agents.
  For $H_2$ we also show the robot's running belief $\distr(M^{H_2} \mid \ivec_{t})$.
  }
\end{figure*}

The performance-safety trade-off curve plotted in Figure~\ref{fig:Ex1_tradeoff} validates H1.
Here, we design each \emph{unshielded} planner with a set of fine-tuned safety-critical parameters, e.g. weight of the soft-constrained collision avoidance cost, robust margin of the failure set, and acceleration limits.
For a given planner design, we simulate the scenario 50 times, each with a different random seed, which affects uncertainty sources including the initial conditions, additive disturbances, human's lane preference, and safety awareness.
These random variables are independent of the (closed-loop) interactions between the human and the robot.
Note that even the least conservative IDSMPC policy still leads to a lower collision rate than the baselines, and yields a closed-loop cost similar to those of non-dual policies.
Although the EDSMPC policy also manages to achieve a low collision rate thanks to its ability to actively reduce human uncertainty, its overall closed-loop performance is consistently inferior to that of IDSMPC, which validates H2.
Finally, we test IDSMPC-SHARP under the 50 scenarios governed by the same random seed.
Thanks to shielding, safety is assured for all 50 trials and, due to shielding awareness, the (normalized) average closed-loop cost ($1.30$) achieved by IDSMPC-SHARP is reduced by about $19\%~(\pm 5.6\%~\text{s.d.})$ compared to that of IDSMPC ($1.61$), which is fine-tuned to achieve zero collision rate.
This validates H3.

Trajectory snapshots and evolution of $\prob(M=\text{l})$ of one simulation trial are shown in Figure~\ref{fig:Ex1_Trajs}.
The ground truth human's lane preference is the right lane for the first $3$~s\footnote{All time data describing agent behaviors in Section~\ref{sec:sim} are simulated.} and then becomes the left lane for the remainder of the simulation, as shown at the bottom of Figure~\ref{fig:Ex1_Trajs}.
The priors are chosen as $P(M=\text{l})=1$ and $\theta^{\text{l}},\theta^{\text{r}} \sim \gaussian((0.5,0.5),5I)$.
Unlike non-dual control planners, IDSMPC controlled the robot to approach the human-driven vehicle along the center of the road, allowing the robot to informatively probe the human---which resulted in a more accurate prediction of $M$ (bottom)---and guiding the robot through a region from which collisions can be avoided more easily.
Indeed, as the human's hidden state $M$ switched from $\text{r}$ to $\text{l}$ at $t = 6$~s, the robot using IDSMPC executed a sharp right turn and successfully avoided colliding with the human.
The EDSMPC planner, although effective at reducing the uncertainty at the beginning, failed to recognize that overtaking the human from the right would have resulted in a more efficient trajectory.
All non-dual control planners, \emph{even with replanning}, caused a collision with the human due to insufficient knowledge about $M$.
It is also worth noticing that even if IDSMPC uses NDSMPC solutions for initialization, their closed-loop behaviors are vastly different, which essentially comes from the dual control effect.

In Figure~\ref{fig:Ex1_safety_liveness}, we examine another simulation trial.
Using IDSMPC, the robot was able to safely overtake the human-driven car in $6$ s.
However, using the ISA-iLQ planner, the robot failed to overtake the human within 10 s.
Due to the lack of dual control effort, the robot was stuck behind the human, unaware of the human's willingness to make room for the robot.
In Figure~\ref{fig:Ex1_safety_liveness}, we also display an unsafe trajectory generated with the CEMPC planner.
Those results demonstrate that, with the dual control effort, the robot gains better safety \emph{and} liveness properties when interacting with the other agent.

\subsubsection{Behavioral and Cooperative Uncertainty.}
\label{sec:sim:Ex2}
We next consider the uncontrolled traffic intersection scenario with the human uncertainty introduced in Example~\ref{example:2}.
Trajectory snapshots of four simulation trials using IDSMPC with different hidden states are shown in Figure~\ref{fig:Ex2}.
We chose uninformative prior distributions $\distr(M \mid \ivec_0)=\begin{bmatrix} 0.25 & 0.25 & 0.25 & 0.25 \end{bmatrix}$ and $\theta^{M} \sim \gaussian((0.5,0.5),5I)$ for all $M \in \mset$.
We see that all four trials were safe and both the autonomous vehicle and pedestrian reached their target.
The performance-safety trade-off curve is plotted in Figure~\ref{fig:Ex1_tradeoff} obtained based on 50 simulated trials, similar to Section~\ref{sec:sim:Ex1}.
Again, we see that the IDSMPC policy leads to the best performance-safety trade-off among all unshielded planners, and a collision rate consistently lower than $15\%$.
Safety is achieved for all trials with the shielded IDSMPC-SHARP policy, resulting in a (normalized) average closed-loop cost ($1.25$), which is $25\%~(\pm 3\%~\text{s.d.})$ lower than what is achieved by the safest IDSMPC design ($1.67$).

\subsubsection{Four-Agent Interaction Example.}
\label{sec:sim:Ex3}
Finally, we apply IDSMPC to the same traffic intersection scenario as in Example~\ref{example:2} involving three human agents: two human-driven vehicles and a pedestrian.
Human-driven vehicles are modeled with the objective and awareness uncertainty (Section~\ref{sec:sim:Ex1}), and the pedestrian is modeled with the behavioral and cooperative uncertainty (Example~\ref{example:2} and Section~\ref{sec:sim:Ex2}).
In addition, we set up the simulation so that the pedestrian ignores other agents ($M = \text{o}$) for the first 2 s and then becomes safety-aware ($M = \text{p}$) for the remainder of the simulation.
Trajectory snapshots of one representative trial are shown in Figure~\ref{fig:Ex3}.
The autonomous vehicle was able to quickly reduce the uncertainty of other agents and safely passed the traffic intersection.

\subsection{Evaluation on the Waymo Motion Dataset}
\label{sec:sim:waymo}
\begin{figure}[!hbtp]
  \centering
  \includegraphics[width=1.0\columnwidth]{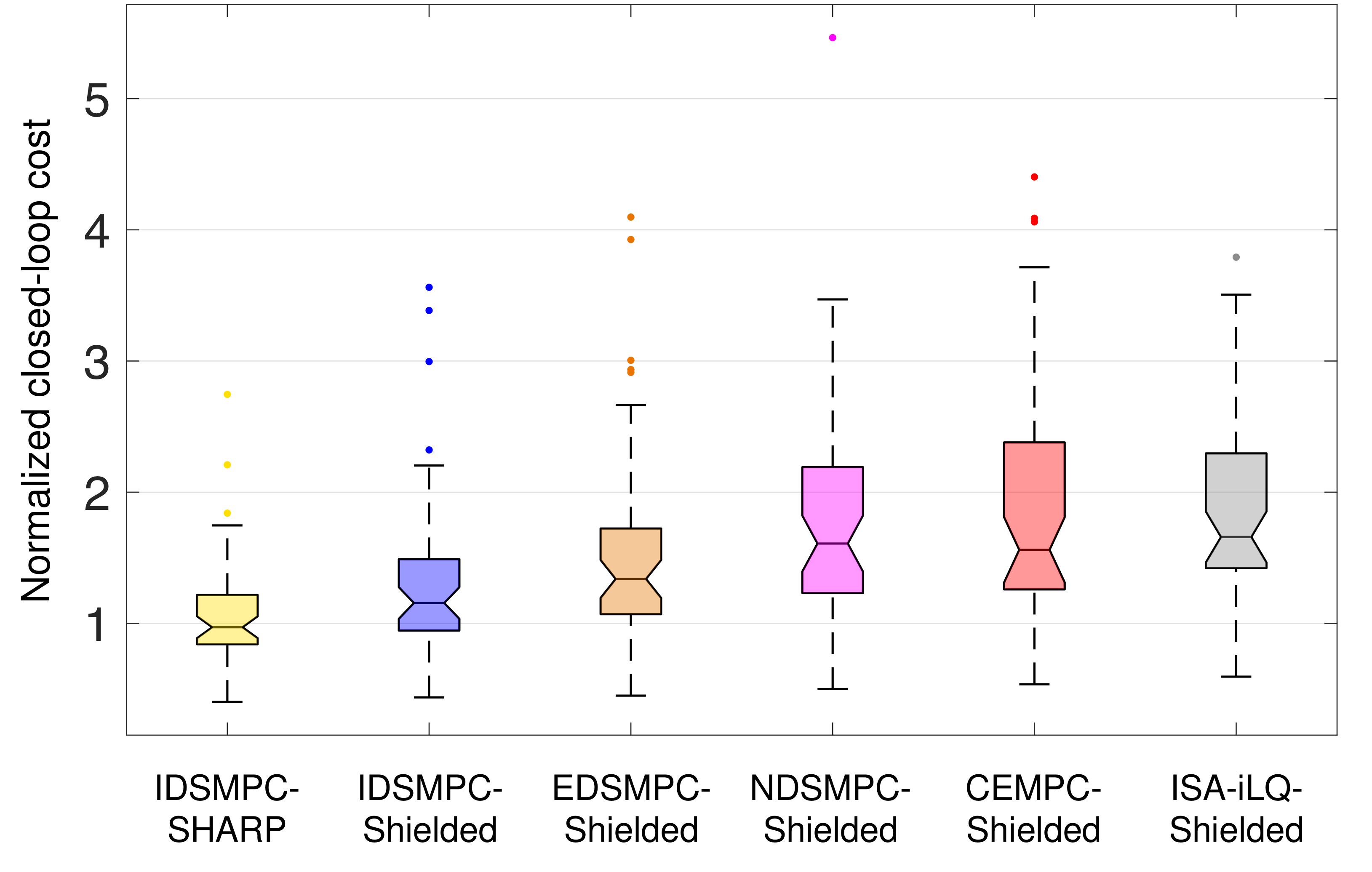}
  \caption{\label{fig:waymo} Closed-loop cost (normalized by $1 \times 10^4$) of the highway driving scenario (Section~\ref{sec:sim:Ex1}) with 50 realistic human-driven vehicle trajectories selected from the Waymo Open Motion Dataset~\cite{sun2020scalability}.
  Central marks, bottom, and top edges of the boxes indicate the median, 25th, and 75th percentiles, respectively.
  The maximum whisker length is set to 1.5, which leads to 99.3\% coverage if the data are normally distributed.
  Outliers are shown as points.}
\end{figure}

In this section, we provide additional simulation results for the highway driving scenario (Section~\ref{sec:sim:Ex1}), where the human driver's trajectories are taken from the Waymo Open Motion Dataset~\cite{sun2020scalability}.
We filtered out 50 trajectory data with different human motions and target lanes from the original dataset.
Since the motion of the human-driven vehicle is generated by replaying the trajectory data, the human can be seen as completely unaware of safety, which is unknown to the robot.
Statistical data of the closed-loop costs (normalized by $1 \times 10^4$) obtained from 50 trials are plotted in Fig.~\ref{fig:waymo}, where all planners are shielded and thus all trials are safe.
The (normalized) average closed-loop cost of IDSMPC-SHARP, IDSMPC, EDSMPC, NDSMPC, CEMPC, and ISA-iLQ are $1.09$, $1.35$, $1.57$, $1.81$, $1.89$, and $1.86$, respectively, as indicated by the central marks of the boxes in Fig.~\ref{fig:waymo}.
Even if the human is unresponsive, dual control planners are still more efficient than non-dual ones due to active uncertainty reduction.
IDSMPC-SHARP outperforms all other planners, showing its applicability under realistic interaction scenarios.

\subsection{Test of Statistical Significance}
\vspace{-0.2cm}
To verify that our results are statistically significant to confirm the three hypotheses made in Section~\ref{subsec:hypotheses}, we performed the analysis of variance (ANOVA) test for results reported in Section~\ref{sec:sim:Ex1},~\ref{sec:sim:Ex2}, and~\ref{sec:sim:waymo}, with the closed-loop cost as the responsive variable, and planning methods grouped pairwise as the independent variable.
First, we found a significant main effect between non-dual SMPC (NDSMPC) and implicit dual SMPC (IDSMPC) for data reported in Section~\ref{sec:sim:Ex1} ($F(1,92)=38.18,~p<0.001$), Section~\ref{sec:sim:Ex2} ($F(1,98)=97.74,~p<0.001$), and Section~\ref{sec:sim:waymo} ($F(1,98)=9.01,~p=0.003$), which validates H1.
Next, by comparing explicit dual SMPC (EDSMPC) to implicit dual SMPC (IDSMPC), we again found a non-negligible effect for data reported in Section~\ref{sec:sim:Ex1} ($F(1,98)=75.49,~p<0.001$) and Section~\ref{sec:sim:Ex2} ($F(1,98)=74.34,~p<0.001$), which supports H2.
Finally, we also found a significant main effect between implicit dual SMPC (IDSMPC) with and without shielding awareness for data reported in Section~\ref{sec:sim:Ex1} ($F(1,98)=39.08,~p<0.001$) and Section~\ref{sec:sim:Ex2} ($F(1,98)=110.63,~p<0.001$), which supports H3.
In the ANOVA tests for results obtained in Section~\ref{sec:sim:waymo}, we noticed a decrease in the F-value and an increase in the p-value when we attempted to validate H2 ($F(1,98)=2.43,~p=0.123$) and H3 ($F(1,98)=5.24,~p=0.024$), which, admittedly, falls short of the statistical significance.
We note that the result is expected since the behavior of the other agent was replayed based on the groundtruth trajectory data from the Waymo Open Motion Dataset, and therefore the other agent is non-responsive to the ego.
Consequently, both the dual SMPC and shielding awareness, which rely heavily on reasoning the interaction among agents, are inevitably less effective.

\section{Hardware Demonstration}
\label{sec:hardware}

\subsection{Experiment Setup}
\begin{figure}[!hbtp]
  \centering
  \includegraphics[width=0.5\columnwidth]{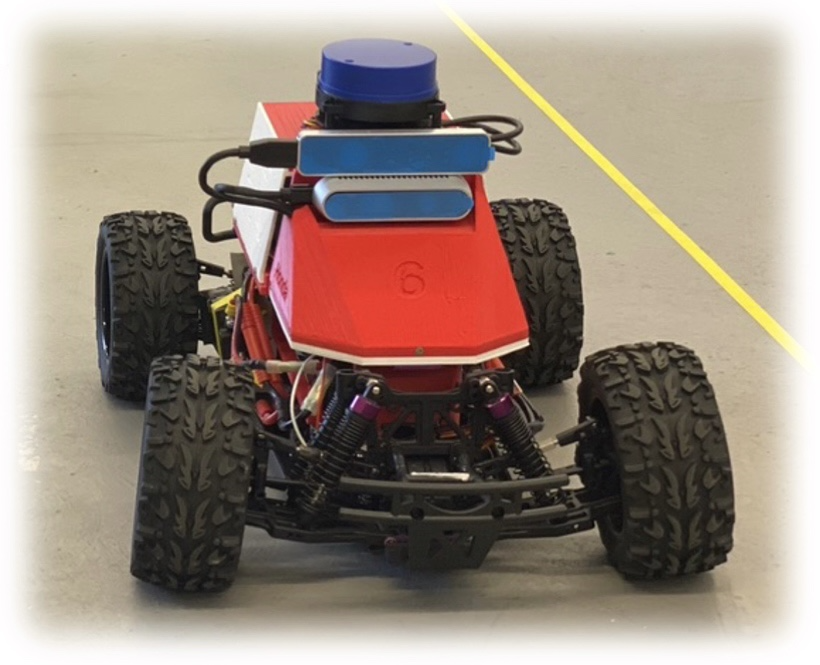}
  \caption{\label{fig:MuSHR} The MuSHR autonomous vehicle equipped with a LiDAR for localization.
  }
\end{figure}

In this section, we demonstrate our proposed IDSMPC-SHARP planning framework (Section~\ref{sec:SHARP}) on Example~\ref{example:1} (overtaking) with two customized 1/10 scale Multi-agent System for non-Holonomic Racing (MuSHR)~\cite{srinivasa2019mushr} autonomous vehicles (one ego and one peer vehicle) at a test track of Honda Research Institute USA, Inc. in San Jose, CA.
The 2D map of the track can be found in Figure~\ref{fig:hardware}.
We use the Robot Operating System (ROS) to establish communications among sensors, actuators, and computing units.
Each MuSHR robot (Figure~\ref{fig:MuSHR}) uses a LiDAR for determining its own state (position, velocity, and orientation) based on a given grid map of the track and surrounding landmarks, and communicates its \emph{current} state with the other vehicle.
The IDSMPC-SHARP and all comparative planners run at 10Hz\footnote{Our code implementation yields a computation time lower than $100$~ms per planning cycle for all control policies.} on an Intel Xeon desktop with an E5-2640 CPU and send to the MuSHR vehicle the planned trajectory, which is tracked by a PID-based low-level controller running on an Intel NUC mini PC onboard MuSHR.
Both the desktop and NUC run the Ubuntu 20.04 LTS operating system.
We use the 4D kinematic bicycle model in~\cite{zhang2020optimization} as the vehicle dynamics in both the ego's and other agent's MPC problem.
All MPC problems in this section are modeled as nonlinear programs (NLPs) with CasADi~\cite{andersson2019casadi} in Python and solved in real time using IPOPT~\cite{wachter2006implementation} with the linear system solving subroutine MA57~\cite{duff2004ma57}.
We used HJ Reachability~\cite{bansal2017hamilton, leung2020infusing} to synthesize the shielding mechanism.
The HJ-based shielding policy is pre-computed with OptimizedDP~\cite{bui2022optimizeddp} and deployed onboard Intel NUC.  
Parameter values used for hardware experiments can be found in Table~\ref{tab:planner_param_hard} in Appendix~\ref{apdx:param:hardware}.

\subsubsection{Comparative Methods.}
We compare our proposed IDSMPC-SHARP planner against four other planners in hardware experiments:
\begin{itemize}
    \item \textbf{Oracle:} An MPC planner that computes the policy based on the other agent's \emph{communicated plans} computed at the current time.
    
    \item \textbf{Ablation~\upperRomannumeral{1}:} IDSMPC, which removes the shielding-aware constraint~\eqref{eq:cvx_sh_constr} from the SMPC problem~\eqref{eq:ST-SMPC}, making it shielding-agnostic.
    
    \item \textbf{Ablation~\upperRomannumeral{2}:} NDSMPC-SHARP, which removes the belief dynamics (measurement update)~\eqref{eq:ST-SMPC:b_dyn} and only updates the belief states with transition model~\eqref{eq:ST-SMPC:b_dyn_trans}.
    The policy therefore passively reduces the uncertainty.
    
    \item \textbf{Baseline:} CEMPC-SHARP, which integrates the CEMPC planner introduced in Section~\ref{sec:sim:setup} with the shielding-aware constraint~\eqref{eq:cvx_sh_constr}.
\end{itemize}
Here, only the IDSMPC policy produces dual control effect among all four comparative methods.
All planners use the same quadratic cost functions $\ell$ and $\ell_F$, and are equipped with the same HJ-Reachability-based shielding mechanism.

\subsubsection{Safety Regulations and Performance Metrics.}

Due to the relatively large size of the vehicle (approx. $40$cm in width) compared to the lane width (approx. $60$cm), we thereby define a trial to be \emph{safe} if both of the following two regulations are satisfied:
\begin{itemize}
    \item \textbf{SR1:} The ego and other vehicles do not collide (i.e. in physical contact) with each other, and
    \item \textbf{SR2:} At least one wheel of the ego vehicle is touching or inside the track limit (shown as the outer grey lines in Figure~\ref{fig:hardware}).
\end{itemize}
This safety regulation is motivated by typical car races, in which, similar to our case, vehicles are allowed to make aggressive overtaking maneuvers on relatively narrow lanes.

Motivated by~\cite{leung2020infusing}, we consider the following two metrics that evaluate the trade-off between safety and efficiency:
\begin{itemize}
    \item Safety index, defined as
    $$SI := \sum_{k=0}^{T_{\text{exp}}} \mathbf{1}\left[\ell_{\text{HJ}}\left(x_k\right) \geq 0\right] \ell_{\text{HJ}}\left(x_k\right),$$
    where $T_{\text{exp}} > 0$ is the total number of time steps of an experiment trial and $\ell_{\text{HJ}}(\cdot)$ is the running cost (signed distance function~\cite{bansal2017hamilton}) used in HJ Reachability computation, which captures the penalty of violating safety regulations SR1 and SR2.
    This index captures the time-accumulated severity of safety violations (with respect to the shielding safe set $\Omega$).
    
    \item Efficiency index, defined as
    \begin{align*}
       EI := &\frac{1}{T_{\text{exp}}} \sum_{k=0}^{T_{\text{exp}}} \|x_k^\ego-x_{k,\text{ref}}^\ego\|^2_{Q_{\text{exp}}} + \|u_k^\ego\|^2_{R_{\text{exp}}} \\
       &+\left(p_{y, k}^\oppo-p_{y, k}^\ego\right), 
    \end{align*}
    where $Q_{\text{exp}}$ and $R_{\text{exp}}$ are cost function matrices used by all five MPC running cost $\ell(\cdot)$, whose values can be found in Table~\ref{tab:planner_param_hard}, and the term $(p_{y, k}^\oppo-p_{y, k}^\ego)$ incurs a penalizing cost when the ego has not overtaken the other agent, and produces a reward otherwise.
    This index captures the time-average planning efficiency measured by a combination of tracking accuracy, control magnitude, and overtaking progress.
\end{itemize}

\subsubsection{Agent's Policy.}
\label{sec:hardware:oppo_policy}
The other agent uses an MPC policy equipped with the shielding-aware constraint~\eqref{eq:cvx_sh_constr} to track the reference lane while avoiding colliding with the ego vehicle.
The soft constraint cost weights of constraint~\eqref{eq:cvx_sh_constr} are randomized across different trials to diversify the other agent's commitment to safety.
When the ego vehicle is within a detection circle of radius $r^\oppo_{\text{detect}}$, the other agent yields to the ego by changing to the other lane with a fixed probability after a time delay.
Both the yielding probability and time delay are randomized across different trials.
Note that the interactive behaviors produced by the other agent's (randomized) shielding-aware MPC policy are different from, and do not use any quantity computed in the motion prediction model defined in Example~\ref{example:1} and used in the ST-SMPC formulation~\eqref{eq:ST-SMPC}.
Therefore, our experiments are not self-fulfilling.

\subsection{Experiment Results}
We start by presenting one set of representative trials (one for each planner) in Figure~\ref{fig:hardware}.
For a fair comparison, we chose the same policy parameters of the other agent across all five trials.
The ego vehicle's reference lane is set to the inner lane for all trials.
With both active uncertainty reduction and shielding-aware robust planning equipped, our proposed IDSMPC-SHARP planner produced a safe and efficient trajectory, whose quality is comparable to the one given by the Oracle MPC policy.
By contrast, due to the lack of shielding awareness, IDSMPC (Ablation~\upperRomannumeral{1}) produced an unsmooth and wobbling trajectory caused by triggering an emergency shielding maneuver when the ego vehicle made a hard left turn to avoid the other agent at close proximity.
Since the belief state dynamics (measurement update) is removed from NDSMPC-SHARP (Ablation~\upperRomannumeral{2}), it only passively learns the values of other agents' hidden states.
As a result, the ego vehicle became overly conservative and was not able to overtake the other agent.
CEMPC-SHARP (Baseline) makes decisions only based on the MAP estimate of hidden states, which oftentimes lags behind the other agent's actual motion.
Indeed, the ego vehicle made a right turn to try to overtake the other agent from the outer lane when the MAP estimated hidden state was $\text{NY}$ (not yielding), but the other agent showed a clear intention to make way for the ego.
The right turn was therefore unnecessary and deemed inefficient for the overall planning performance. 

\begin{figure}[!hbtp]
     \centering
     \includegraphics[width=1.0\columnwidth]{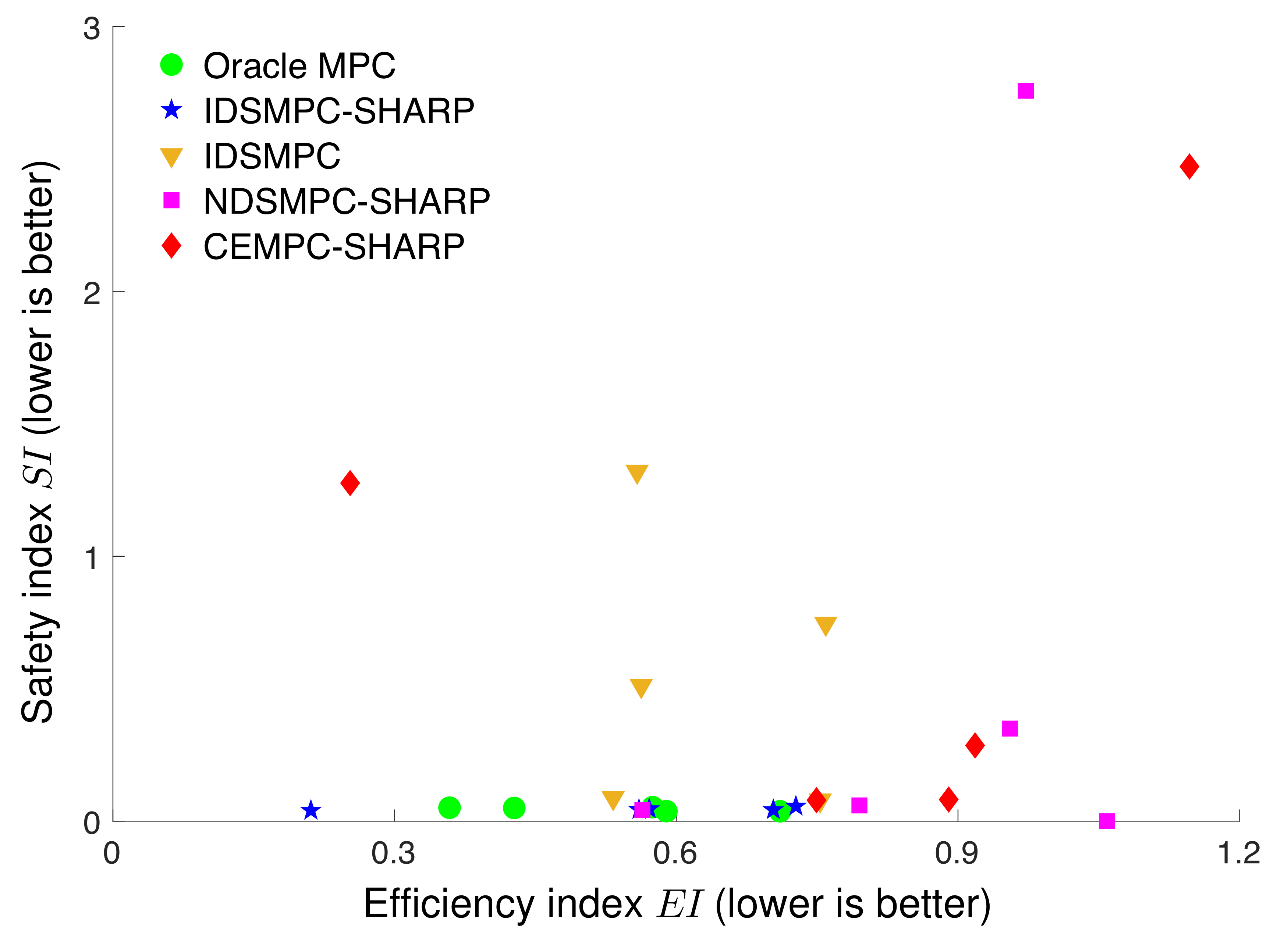}
     \caption{\label{fig:tradeoff_hardware} Performance and safety trade-off of hardware experiments (Example 1).
     Each data point is obtained based on the closed-loop trajectory of one trial.
     The lower left corner is the desired region, where both $EI$ and $SI$ are low.
     }
\end{figure}

Next, we performed a performance-safety trade-off study using the hardware experiments, and the results are plotted in Figure~\ref{fig:tradeoff_hardware}.
Here, we ran 5 trials for each planner, each with a different random seed, which affects the initial conditions and the other agent's policy parameters.
\emph{All trials were safe according to safety regulations SR1 and SR2.}
Note that in order to account for model mismatch and communication delays, we designed a more conservative HJ Reachability shielding mechanism than SR1 and SR2, resulting in safety index $SI > 0$ for almost all trials.
We see that IDSMPC-SHARP maintained a good balance between safety and efficiency, similar to that of the Oracle MPC.
It is worth noticing that even the most conservative NDSMPC-SHARP policy can lead to poor safety performance.
This is because, without active uncertainty reduction, SHARP was not able to effectively predict future shielding events given belief states with high uncertainty, and ultimately led the system to enter a near-unsafe region, incurring a large safety index $SI$.
CEMPC-SHARP, due to strategy alignment, is sensitive (less robust) to belief fluctuations caused by the randomness of the experiments.
Therefore, its data points have the highest variance among all five planners.

\begin{figure*}[!hbtp]
\centering
\subfloat[Hardware experiment trial of Example~\ref{example:1} using Oracle MPC. Plotted trajectories are subject to minor LiDAR localization error.]{
    \label{fig:hardware:OMPC}
    \includegraphics[width=2.0\columnwidth]{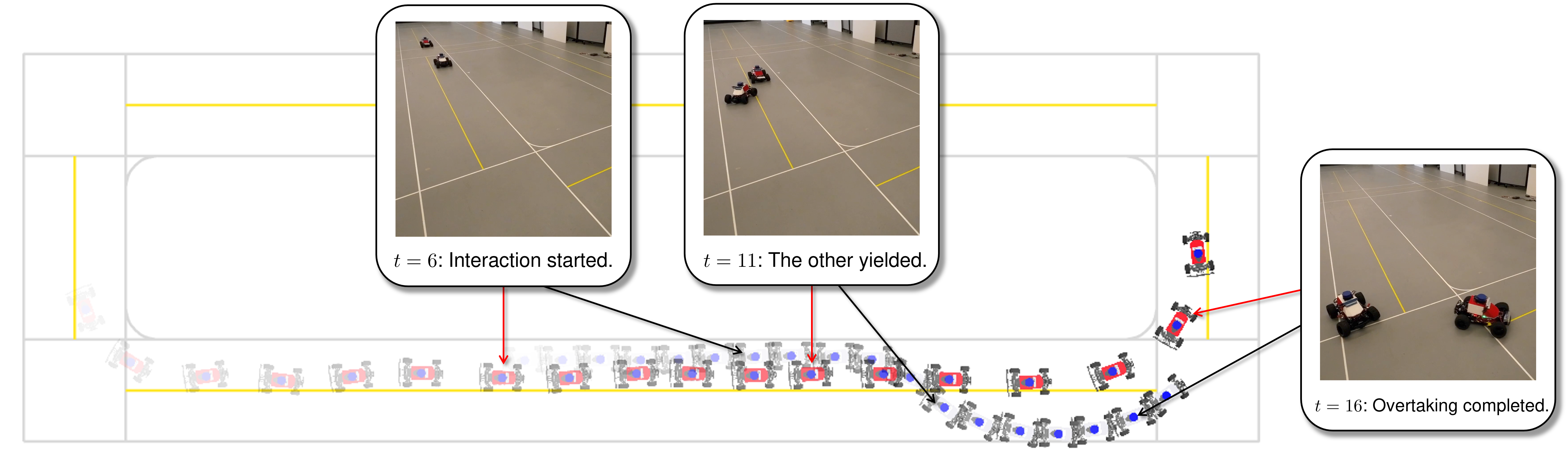}
}

\subfloat[Hardware experiment trial of Example~\ref{example:1} using IDSMPC-SHARP (proposed). Plotted trajectories are subject to minor LiDAR localization error.]{
    \label{fig:hardware:IDSMPC-SHARP}
    \includegraphics[width=2.0\columnwidth]{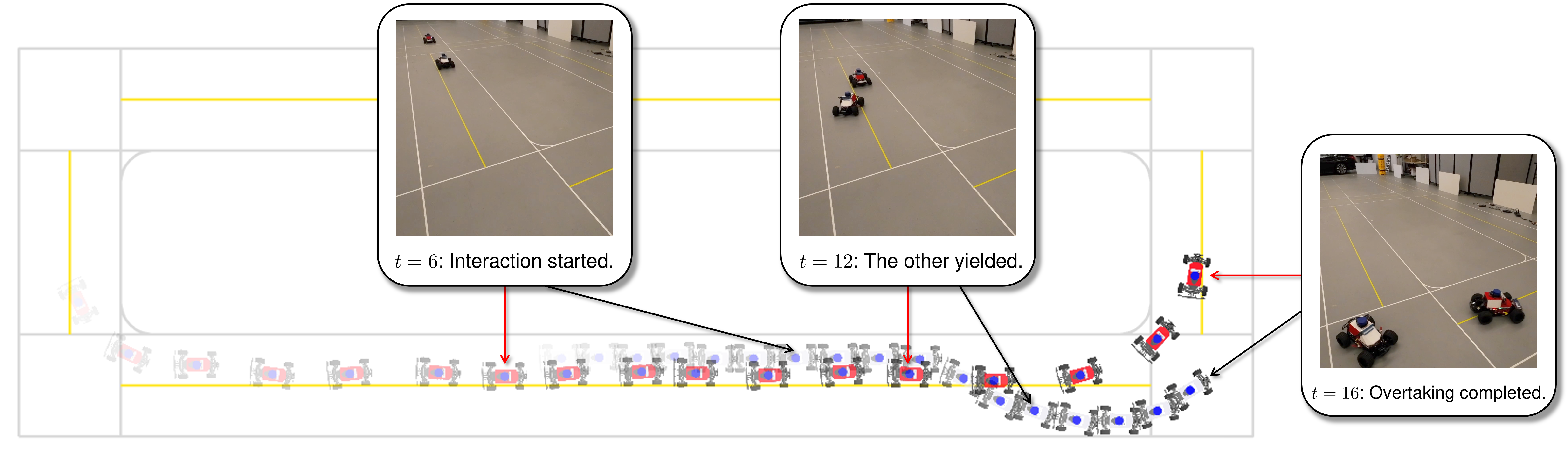}
}

\subfloat[Hardware experiment trial of Example~\ref{example:1} using IDSMPC. Plotted trajectories are subject to minor LiDAR localization error.]{
    \label{fig:hardware:IDSMPC}
    \includegraphics[width=2.0\columnwidth]{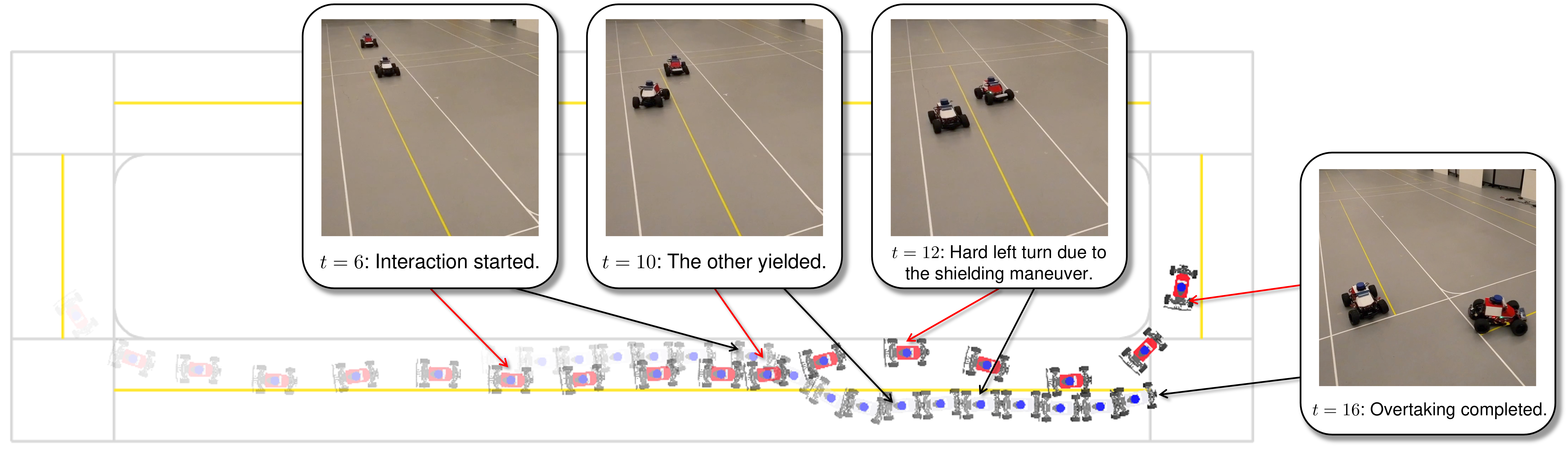}
}

\subfloat[Hardware experiment trial of Example~\ref{example:1} using NDSMPC-SHARP. Plotted trajectories are subject to minor LiDAR localization error.]{
    \label{fig:hardware:NDSMPC-SHARP}
    \includegraphics[width=2.0\columnwidth]{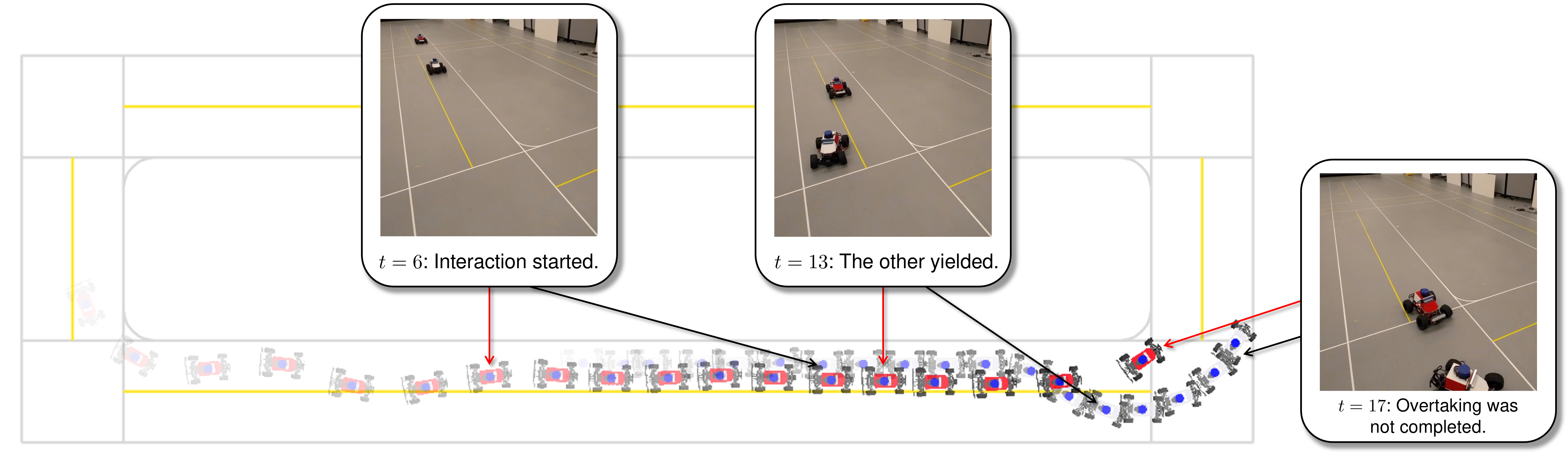}
}
\caption{Comparison between the proposed IDSMPC-SHARP planner and four other planning strategies.
The ego and other vehicles are painted with a red and white front, respectively.
Vehicle snapshots are plotted every 1 second.
The proposed IDSMPC-SHARP planner accurately predicted the other agent's willingness to yield and performed a clean overtaking maneuver, comparable to the motion produced by the Oracle MPC, which computes the policy based on the other agent's communicated future plans.
The IDSMPC planner (without shielding awareness) suffered performance loss due to an unforeseen emergency shielding maneuver.
Without active uncertainty reduction, NDSMPC-SHARP produced an overly conservative motion and was unable to complete the overtaking task.
CEMPC-SHARP is sensitive to errors in the MAP estimate of the other agent's hidden states and resulted in an unnecessary lane changing maneuver due to a lag in identifying the other agent's willingness to yield. Video can be found at \url{https://youtu.be/sJawMtO5QgY}.
\label{fig:hardware}}
\end{figure*}
\begin{figure*}[!hbtp]\ContinuedFloat
  \centering
  \subfloat[Hardware experiment trial of Example~\ref{example:1} using CEMPC-SHARP. Plotted trajectories are subject to minor LiDAR localization error.]{
    \label{fig:hardware:CEMPC}
    \includegraphics[width=2.0\columnwidth]{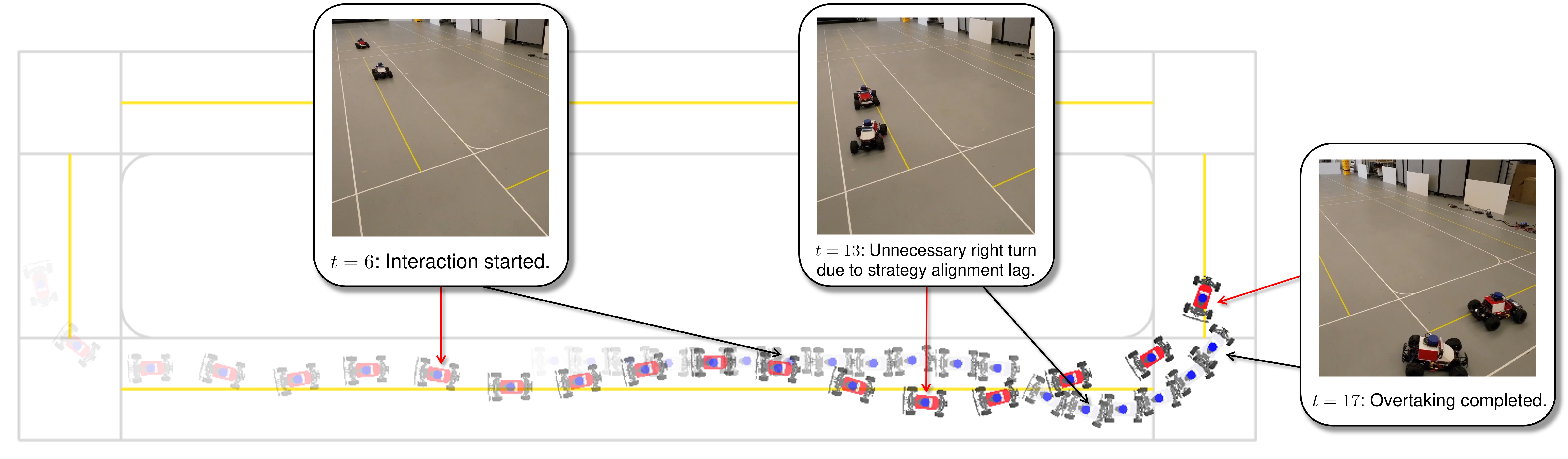}
}
\caption{Comparison between the proposed IDSMPC-SHARP planner and four other planning strategies (cont'd).
}
\end{figure*}

\section{Conclusions}

We have introduced an implicit dual control approach towards active uncertainty reduction for interaction planning.
The resulting policy improves planning efficiency via a tractable approximation to the Bellman recursion of a dual control problem, leading to an implicit dual scenario-based SMPC policy, which automatically achieves an efficient balance between optimizing expected performance and eliciting information on future behaviors of the other agent.

Robust safety guarantee is obtained by wrapping the dual control policy with shielding, a supervisory safety filter.
The SMPC problem is augmented with a convex shielding-aware constraint derived based on an improved variant of the recently proposed SHARP framework.
The resulting IDSMPC-SHARP policy allows the ego robotic agent to efficiently interact with the other agent, while being aware of the risk of applying the costly shielding maneuvers triggered by unlikely actions of the other agent.
We demonstrate the proposed framework with simulated driving examples and ROS-based hardware experiments using 1/10 scale autonomous vehicles.

\subsection{Limitations}
Although we have demonstrated our method on an interaction planning example with three peer agents, generalizing to more agents remains an open challenge.
In the worst case, the number of nodes (hence decision variables) grows exponentially with the number of interacting agents, and the number of exploration (dual control) time steps.
Still, the scenario-based MPC approach is suitable for moderate-sized interaction planning problems.
In addition, the current framework assumes that the ego agent can perfectly observe other agents' state and past actions, which is often unrealistic.
Recent advances in interaction planning with observation uncertainties~\cite{isele2018navigating, sunberg2022improving,hu2023belgame} provide a promising direction to improve and generalize our method in such settings.

\subsection{Future Directions}
We see our work as an important step towards a broader class of methods that handle different parametrizations of other agents' behavior from the Boltzmann rationality model, including the quantal level-$k$ model~\cite{stahl1994experimental,tian2021anytime}, learning-based prediction~\cite{isele2018safe}, nonlinear opinion dynamics~\cite{bizyaeva2022nonlinear}, and state- and input-dependent belief state transition dynamics $g^t(\bel^-, \state, \ctrl^\ego)$ that captures the effect of ego's decisions on the other agent's hidden state.
While this paper focuses on the robot's own performance, our approach may be adapted to account for social coordination and altruism~\cite{toghi2021social} in cooperative human-robot settings.
We are also excited to test our framework on other interaction planning tasks such as human-drone interaction~\cite{fisacBHFWTD18} with real human participants.

\begin{acks}
This work is supported by the Princeton SEAS Project X Innovation Fund and the Honda Research Institute (HRI) USA, Inc. This article solely reflects the opinions and conclusions of its authors and not HRI, or any other Honda entity.
The authors thank Thang Lian, Huan D. Nguyen, and Zhaobo K. Zheng for their help with the hardware experiments.
The authors also thank Faizan M. Tariq, Piyush Gupta, Aolin Xu, Yichen Song, Zixu Zhang, and Kai-Chieh Hsu for very helpful discussions on decision making under uncertainty, MPC, and shielding.
\end{acks}

\bibliographystyle{SageH}
\bibliography{reference}

\nomenclature[A]{$\ego$}{Ego}
\nomenclature[A]{$\oppo$}{Other agent}
\nomenclature[A]{$\state$}{Physical state}
\nomenclature[A]{$\bel$}{Belief state}
\nomenclature[A]{$\ctrl$}{Control input}
\nomenclature[A]{$\dyn$}{Autonomous part of joint dynamics}
\nomenclature[A]{$\beldyn$}{Belief state dynamics}
\nomenclature[A]{$\beldyn^t$}{Belief transition dynamics}
\nomenclature[A]{$\beldynapprox$}{Approximate belief state dynamics}
\nomenclature[A]{$\dstb$}{External disturbance input}
\nomenclature[A]{$\theta$}{Continuous hidden state}
\nomenclature[A]{$M$}{Discrete hidden state (mode)}
\nomenclature[A]{$\qfunc$}{State-action value (Q-value) function}
\nomenclature[A]{$\ell$}{Ego's stage cost function}
\nomenclature[A]{$\ell_F$}{Ego's terminal cost function}
\nomenclature[A]{$\ivec$}{Information vector}
\nomenclature[A]{$F$}{System matrix in parameter-affine dynamics}
\nomenclature[A]{$\dstbbar$}{Combined disturbance in parameter-affine dynamics}

\nomenclature[B]{$\node$}{Node}
\nomenclature[B]{$\node_0$}{Root node}
\nomenclature[B]{$\pre{\node}$}{Parent node of node $\node$}
\nomenclature[B]{$N$}{Planning horizon}
\nomenclature[B]{$N^d$}{Dual control horizon}
\nomenclature[B]{$N^e$}{Exploitation horizon}
\nomenclature[B]{$\nodeset$}{Node set}
\nomenclature[B]{$\nodeset^d$}{Dual control node set}
\nomenclature[B]{$\nodeset^e$}{Exploitation node set}
\nomenclature[B]{$\nodeset^s$}{Shielding node set}
\nomenclature[B]{$\nodesetleaf$}{Leaf node set}
\nomenclature[B]{$\probbar_\node$}{Transition probability of node $\node$ from its parent}
\nomenclature[B]{$\prob_\node$}{Path transition probability of node $\node$}

\nomenclature[C]{$\cset$}{Control bound}
\nomenclature[C]{$\dset$}{Disturbance bound}
\nomenclature[C]{$\failure$}{Failure set}
\nomenclature[C]{$\Omega$}{Safe (controlled-invariant) set}
\nomenclature[C]{$\Omega_\node$}{Approximate local safe set at node $\node$}
\nomenclature[C]{$\policy^\shield$}{Shielding policy}
\nomenclature[C]{$\policy^\sfilter$}{Safety filter policy}
\nomenclature[C]{$h$}{Robust control barrier function used in SHARP}

\printnomenclature

\appendix

\section{Parameter Values}
\label{apdx:param}

\subsection{Parameter Values used in Section~\ref{sec:sim}}
\label{apdx:param:sim}

\begin{table}[H]
    \caption{Planner and simulation parameters used in Section~\ref{sec:sim}.}
    \label{tab:planner_param_sim}
    \centering
    \begin{tabular}{|c|c|p{0.21\textwidth}|}
    \hline
    \textbf{Notation} & \textbf{Value} & \textbf{Definition} \\
    \hline
    \hline
        $\Delta t$ & 0.2 s &  Sampling time \\
        $l_w$ & 3.7 m & Lane width \\
        $Q_{\text{sim}}$ & $\diag(1,2,1,1)$ & Ego's state cost matrix \\
        $R_{\text{sim}}$ & $\diag(0.1,1)$ & Ego's control cost matrix \\
        $\Sigma^d$ & $0.1 I$ & Disturbance covariance\\
        $N^d$ & 2 & Dual control time steps\\
        $N^e$ & 4 & Exploitation time steps\\
        $K$ & 2 & Branching number of $\theta^M$\\
        $\gamma$ & 0.5 & RCBF constraint parameter used in SHARP\\
    \hline
    \end{tabular}
\end{table}

\subsection{Parameter Values used in Section~\ref{sec:hardware}}
\label{apdx:param:hardware}

\begin{table}[H]
    \caption{Planner and experiment parameters used in Section~\ref{sec:hardware}.}
    \label{tab:planner_param_hard}
    \centering
    \begin{tabular}{|c|c|p{0.21\textwidth}|}
    \hline
    \textbf{Notation} & \textbf{Value} & \textbf{Definition} \\
    \hline
    \hline
        $\Delta t$ & 0.2 s &  Sampling time \\
        $l_w$ & 0.6 m & Lane width \\
        $Q_{\text{exp}}$ & $\diag(0.5,2,1,1)$ & Ego's state cost matrix  \\
        $R_{\text{exp}}$ & $\diag(0.1,0.5)$ & Ego's control cost matrix \\
        $\Sigma^d$ & $0.01 I_8$ & Disturbance covariance\\
        $N^d$ & 3 & Dual control time steps\\
        $N^e$ & 15 & Exploitation time steps\\
        $K$ & 2 & Branching number of $\theta^M$\\
        $\gamma$ & 0.5 & RCBF constraint parameter used in SHARP\\
        $r^\oppo_{\text{detect}}$ & 2.5 m & Detection circle radius of the other agent \\
        $\bar{t}^\oppo_d$ & 2.5 s &  Average reaction time of the other agent \\
    \hline
    \end{tabular}
\end{table}

\section{Proofs}
\label{apdx:proof}

\subsection{Proof of Lemma~\ref{lem:theta_update}}
\label{apdx:proof:theta_update}
Recall that the combined disturbance term $\bar{\dstb}_t$ in~\eqref{eq:param_affine_dyn} is a zero-mean Gaussian random variable whose covariance is defined in~\eqref{eq:combined_dstb_covar}.
Conditioned on $\state_t \in \ivec_t$, $\ctrl^\ego_t$, and $\theta^M$, the state distribution of $\state_{t+1}$ (likelihood) is Gaussian-distributed, i.e. $\distr(\state_{t+1} \mid \ctrl^\ego_t, \ivec_t; \theta^M, M) \sim  \gaussian\left(\mean^\state_{t+1}, \covar^\state_{t+1}\right)$ whose mean and covariance are given by
\begin{equation*}
\begin{aligned}
\label{eq:state_distr}
\mean^\state_{t+1} &= F(\state_t, \ctrl^\ego_t) \theta^M + \bar{\dyn} (\state_t, u^\ego_t), \\
\covar^\state_{t+1}  &= \covar^\dstb + B^\oppo \covar_t^{\ctrl^\oppo}(\state_t, \ctrl^\ego_t; \theta^M){B^\oppo}^\top =\covar^{\bar{\dstb}}_t.
\end{aligned}
\end{equation*}
Since dynamics~\eqref{eq:param_affine_dyn} are affine in parameter $\theta^M$, applying the self-conjugate property of Gaussian distributions yields the expression of $\mean^{\theta^M_{-}}_{t+1}$ and $\covar^{\theta^M_{-}}_{t+1}$.
\qed

\subsection{Proof of Lemma~\ref{lem:CBF}}
\label{apdx:proof:CBF}
Let $h(\dstate) = H_\tnode^\top \dstate$.
Without loss of generality, let $\dstate_0 =  \state_0 - \bstate_0 = 0$, which implies that $h(\dstate_0) = 0$.
Plugging~\eqref{eq:linearized_sys} into Condition 2 in Definition~\ref{def:RCBF} gives:
\begin{equation*}
\begin{aligned}
    &\forall \tilde{d} \in \dset: \\
    &\quad H_{\tnode}^\top \left[ \left(A_{\tnode} + (\gamma-1)I \right) \dstate + B^\ego_{\tnode} \ctrl^\ego + B^\oppo_{\tnode} \ctrl^\oppo + \tilde{d} \right] \geq 0,
\end{aligned}
\end{equation*}
which is satisfied if there exists $\ctrl^\ego \in \cset^\ego$ such that:
\begin{equation*}
\begin{aligned}
    \min_{\tilde{d} \in \dset} H_{\tnode}^\top \left[ \left(A_{\tnode} + (\gamma-1)I \right) \dstate + B^\ego_{\tnode} \ctrl^\ego + B^\oppo_{\tnode} \ctrl^\oppo + \tilde{d} \right] \geq 0,
\end{aligned}
\end{equation*}
where the optimal disturbance $d^*$ is given by~\eqref{eq:cvx_sh_constr:opt_dstb}.
Therefore, map $h(\dstate)$ is a valid discrete-time Exponential RCBF for system~\eqref{eq:linearized_sys} and safe set $\Omega_\node$.
Using Proposition 4 in~\cite{agrawal2017discrete} we conclude that $\Omega_\tnode$ is robust controlled-invariant.
\qed

\section{Practical Aspects}
\label{sec:practical}

\subsection{Computing Agent's Rational Action}
\label{sec:practical:rational}
The Laplace approximation used by~\eqref{eq:Laplace} requires the mean function (human's rational action) $\mu_i^M(\state_t, \ctrl^\ego_t)$ as the maximizer of the basis Q-value function $\qfunc_{i}^{M}\left(\ctrl_{i}^{M,\oppo}; \state_{t}, \ctrl^\ego_t\right)$.
In our paper, we use the game-theoretic approach~\cite{fridovich2020efficient} as the backend to compute $\qfunc_{i}^{M}(\cdot)$ online, which adopts an analytical maximizer.
In case when the expression of $\mu_i^M(\cdot)$ cannot be computed beforehand, we use a numerical approach similar to~\cite{sadigh2018planning}, which computes a \emph{local} maximizer $\mu_i^M(\cdot)$ during online optimization.
Under the mild assumption that $\qfunc_{i}^{M}(\cdot)$ is a smooth function whose maximum can be attained, we can set the gradient of $\qfunc_{i}^{M}(\cdot)$ with respect to $\ctrl_{i}^{M}$ to $0$.
This condition can be enforced either as a differential-algebraic equation (DAE) constraint~\cite{andersson2019casadi} or a penalty cost in ST-SMPC problem~\eqref{eq:ST-SMPC}.

\subsection{Projecting Predicted Agent's Action}
\label{sec:practical:unbounded}
\looseness=-1
Using the approximate agent's action model~\eqref{eq:exo-agent_ctrl_model_approx} constraint $\ctrl^\oppo_{{\tnode}} \in \cset^\oppo$ in~\eqref{eq:ST-SMPC} may not be feasible, since the predicted human's action $\ctrl^\oppo_{{\tnode}}$
is given as a weighted sum of (unbounded) basis functions with (unbounded) normally distributed weights.
To reconcile this, we define, for each node ${\tnode}$, two separate decision variables: $\tilde{\ctrl}^\oppo_{{\tnode}}$, which must equal the sampled linear combination of basis functions, and $\ctrl^\oppo_{{\tnode}}$, which must satisfy $\ctrl^\oppo_{{\tnode}} \in \cset^\oppo$.
By adding a cost term $C\|\tilde{\ctrl}^\oppo_{{\tnode}}-\ctrl^\oppo_{{\tnode}}\|_2$ to~\eqref{eq:ST-SMPC}, with some large $C>0$ (we use $C=10^8$), the solver sets ${\ctrl}^\oppo_{{\tnode}}$ to the nearest point in $\cset^\oppo$ to the sample-consistent $\tilde{\ctrl}^\oppo_{{\tnode}}$.
This feasible ``projected'' control ${\ctrl}^\oppo_{{\tnode}}$ enters the dynamics in~\eqref{eq:ST-SMPC}.

\subsection{Initialization Pipeline}
\label{sec:practical:init}
\looseness=-1
Since problem~\eqref{eq:ST-SMPC} is in general a large-scale nonconvex optimization problem, initialization is crucial for solving it rapidly and reliably in real time.
In this paper, we generate an initial guess for~\eqref{eq:ST-SMPC} using the following pipeline:
\begin{enumerate}
    \item (Optional) Solve a certainty-equivalent MPC by setting $\theta^M$ and $M$ to their maximum a posteriori estimated values based on the current belief state $\hat{b}_t$.
    \item Solve a non-dual SMPC with the same scenario tree structure as the dual-SMPC, replacing belief state dynamics $\tilde{g}(\cdot)$ with $g^t(\cdot)$ for all dual control steps, and using the certainty-equivalent MPC solution as the initial guess.
    \item Forward-propagate belief states through $\tilde{g}(\cdot)$ using the non-dual SMPC solution for all dual control nodes in the scenario tree.
\end{enumerate}
Step 1 is optional and is only needed when the non-dual SMPC in Step 2 cannot be readily solved.
In Section~\ref{sec:sim} and~\ref{sec:hardware}, we show that even if~\eqref{eq:ST-SMPC} uses results of the non-dual SMPC as its initialization, the resulting closed-loop trajectories are significantly different.

\end{document}